\documentclass[journal]{IEEEtran}

\ifCLASSINFOpdf
   \usepackage[pdftex]{graphicx}
   \DeclareGraphicsExtensions{.pdf,.jpeg,.png}
\else
\fi
\usepackage{amsmath}
\usepackage[linesnumbered,ruled,vlined]{algorithm2e}
\makeatletter
\renewcommand{\@algocf@capt@plain}{above}% formerly {bottom}
\makeatother

\usepackage[usenames, dvipsnames]{color}

\definecolor{green}{rgb}{0,0.5,0}
\definecolor{orange}{rgb}{1,0.4,0}
\definecolor{blue}{rgb}{0.4,0.4,1}
\definecolor{red}{rgb}{1,0,0}

\newcommand{\GG}[1]{{\color{blue} #1}}
\newcommand{\off}[1]{}

\usepackage{array}
\usepackage{calc}

\ifCLASSOPTIONcompsoc
  \usepackage[caption=false,font=normalsize,labelfont=sf,textfont=sf]{subfig}
\else
  \usepackage[caption=false,font=footnotesize]{subfig}
\fi

\usepackage{epstopdf}

\begin{document}

%\title{Non-rigid Affinity Space for Facial Image Quality Enhancement Based on Personal Priors}
\title{Blind Facial Image Quality Enhancement using Non-Rigid Semantic Patches}

\author{Ester~Hait~and~Guy~Gilboa,~\IEEEmembership{Member,~IEEE}% <-this % stops a space
\thanks{E. Hait and G. Gilboa are with the Department of Electrical Engineering,
	Technion - Israel Institute of Technology, Technion City, Haifa, 32000,
	Israel. E-mail: etyhait@campus.technion.ac.il, guy.gilboa@ee.technion.ac.il}% <-this % stops a space
}


\markboth{ieee transactions on image processing,~Vol.~XX, No.~X, August~2016}%
{Shell \MakeLowercase{\textit{et al.}}: Bare Demo of IEEEtran.cls for IEEE Journals}



\maketitle
\begin{abstract}
We propose to combine semantic data and registration algorithms to solve various image processing problems such as denoising, super-resolution and color-correction.
It is shown how such new techniques can achieve significant quality enhancement, both visually and quantitatively, in the case of facial image enhancement. Our model assumes prior high quality data of the person to be processed, but no knowledge of the degradation model.
We try to overcome the classical processing limits by using semantically-aware patches, with adaptive size and location regions of coherent structure and context, as building blocks. The method is demonstrated on the problem of cellular photography enhancement of dark facial images for different identities, expressions and poses.
\end{abstract}

% \fi
\IEEEpeerreviewmaketitle
\section{Introduction}\label{sec:introduction}
\IEEEPARstart
{I}{n} the past decades, handling common image flaws has gradually improved with the use of more sophisticated image priors and models.
%-based priors, describing knowledge of the image space using an image model.
Early methods used pixel-based statistics, such as smoothness~\cite{horn1981determining}, piecewise smoothness~\cite{geman1992constrained}, total-variation~\cite{rudin1992nonlinear}, pixel correlation~\cite{huang1999statistics}, or wavelet decomposition~\cite{simoncelli1999bayesian} for image reconstruction.
In recent years, %Later on,
nonparametric patch-based methods, such as Nonlocal Means~\cite{buades2005review} and BM3D~\cite{dabov2007color}, exploited local and nonlocal self-similarities.
Other patch-based, training-based methods were using Markov Random Fields \cite{roth2005fields} and dictionary learning \cite{aharon2006img}.
Today's main state-of-the-art methods
%{\color{red} Add some recent refs like "On Learning Optimized Reaction Diffusion Processes for Effective Image Restoration", Chen, Pock CVPR 2015;  "Asymptotic Performance of Global Denoising", Talebi-Milanfar SIAM-IS 2016; "Boosting of image denoising algorithms", Romano-Elad.}
are based on square
patches with little if any semantic context \cite{chen2015learning}, \cite{talebi2016asymptotic}, \cite{romano2015boosting}.
%, a sparse image coding method using an adaptive training-based dictionary.
In recent years, using \textit{generic} image priors has started to reach an optimality bound; for example, for super-resolution~\cite{baker2002limits} and denoising~\cite{levin2011natural}. For facial images, \textit{facial priors} were then used to break this limit; For example, face hallucination~\cite{baker2000hallucinating}, or image compression using K-SVD~\cite{bryt2008compression}.
We propose an alternative concept of using large non-rigid patches with high semantic value.

\begin{figure*}[t]
	\captionsetup{justification=centering}
	\centering
	\includegraphics[width=7in]{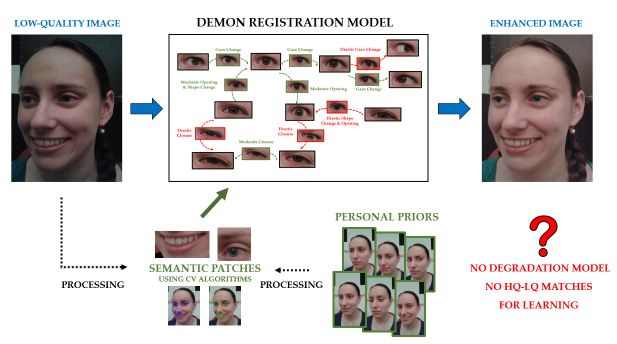}
	\caption{Problem and assumptions of model: blind enhancement of low-quality facial images using prior data. Semantic patches of facial features are extracted to preserve structure and context coherency. Our model assumes available high-quality personal priors, but no knowledge of the degradation model and no matches of HQ and LQ data for learning.}
	\label{fig::model}
\end{figure*}

Fig. \ref{fig::model} demonstrates our model and its underlying assumptions. We aim to use \textcolor{green}{\textit{non-rigid}} processing of \textcolor{green}{\textit{semantic patches}} of facial features, while preserving structure and context coherency, to overcome the classical processing limits.
Given today's highly available mobile photography devices, our model assumes using \textcolor{green}{\textit{high-quality personal priors}} but \textcolor{red}{\textit{no knowledge of the degradation model}}.
The degradation can involve noise following possible nonlinear processing, resolution reduction, a certain degree of motion blur and contrast and color changes.
Our approach suggests to solve the problem indirectly by a mechanism which is invariant to low-to-moderate quality reductions.
  %involving noise and resolution reduction, as well as the camera's built-in image processing.
We also assume that no matches of high quality (HQ) and low quality (LQ) data are available for learning.
As there is no degradation model, one also cannot generate faithfully LQ images by degrading HQ images (e.g. adding noise to a clean image).
Experimental results are demonstrated on the problem of dark cellular image enhancement.
% enhancement for different identities, expressions and poses, using only tens of priors.
%In this work we aim to overcome this limit by using semantically-aware processing to handle regions of coherent structure and context, instead of patches, preserving the structural coherency of highly non-rigid features.
%Specifically, the massive amount of personal images, which will become even more available in the coming decade, due to high-quality digital photography tools, suggests using an external database of same-identity, multiple-expression comprehensive facial priors for facial image enhancement.

\subsection{Related Work}
%{\color{red} THIS PART CAN BE A LITTLE SHORTER (made shorter first 2 paragraphs)}
Capel and Zisserman~\cite{capel2001super} observed that better learning is obtained when considering different facial regions, rather than the whole face, and that better representation is needed when handling high-detail facial regions that attract human attention. A separate PCA basis was learned for different key facial regions.
%; for example, representing the smooth, low-detail cheek required fewer principal components than the eye. Major differences of their work from ours are their
Unlike our proposed method, they use linear PCA decomposition and training sets of \textit{multiple} people.

Jia and Gong~\cite{jia2008generalized} performed face hallucination of a single modality (expression, pose and illumination) into a set of high-resolution images of different modalities, but used multiple people's images as priors. %They formulated a unified hierarchical tensor space representation.
%, which can be reduced into global and local components.
%First, they used a global image-based tensor to synthesize LR images of different modalities from a given LR image. Using a trained local patch-based multiresolution tensor, they incorporated HR details into the cross-modality process to construct a HR image for each modality.
Interestingly, they deduced that hallucinating the same expression as in the test image was better than hallucinating other expressions.
%It is interesting to note the following conclusion made by the authors: for a particular test expression, the hallucination of the same expression was better than the hallucination of other expressions.

Lee \textit{et al}.~‎\cite{lee2003video} represented multiple-pose facial images as a low-dimensional appearance manifold in the image space, for video face recognition. The appearance manifold, learned from training, consisted of pose manifolds and their connectivity matrix, encoding transition probabilities between images.

%Lee \textit{et al}.~‎\cite{lee2003video} represented facial images of different poses as a low-dimensional appearance manifold in the image space, and used it for face recognition in video. The appearance manifold consisted of a collection of pose manifolds and their connectivity matrix, encoding transition probabilities between images, and was learned from a training video.

Yu \textit{et al}.~\cite{yu2007super} incrementally super-resolved 3D facial texture from video under changing light and pose, but used temporal information from sequential frames and a generic 3D face model.
%This was done using Iterative Back Projection; illumination, 3D motion and shape parameters were recovered from tracking.
%They recovered illumination, 3D motion and shape parameters from tracking, and used them to super-resolve 3D texture using Iterative Back Projection.
% As new frames arrived the process iterated, using the super-resolved texture as input to the tracking stage, for improved estimation of illumination and motion parameters.
They also handled facial non-rigidity using a local region-based approach: using a match statistic to detect significant facial regions expression changes between frames.

Shih \textit{et al}~\cite{shih2013joint} performed noise level estimation for denoising, by maximizing the joint noise probability across same-identity facial images of different noise levels.
The estimated noise level can then be used for state-of-the-art denoising algorithms requiring it, such as BM3D.
%In~\cite{shih2013joint}‎, Shih \textit{et al}. compared multiple same-identity aligned facial images, with different noise levels, to jointly estimate an image's noise level for denoising. Noise estimation was based on a probabilistic formulation which aimed to maximize the joint noise probability across images.
%%This was done by first estimating the most probable relative noise levels between image pairs, and then jointly optimizing: modeling using Markov random field, and solving using a fully connected graph.
%The estimated noise level can then be used for state-of-the-art denoising algorithms that require knowledge of the noise level, such as BM3D.

Joshi \textit{et al}.~\cite{joshi2010personal} were the first to suggest the use of "personal priors" to enhance a particular person's image, performing both global and face-specific corrections. They relied on the growing available datasets of personal images. Their algorithm derived its strength from using multiple same-identity example images, which, as they observed, can span a smaller space than that spanned by images of multiple people.

%As a preprocessing stage, they used facial features detection to align facial images and compute a face mask.
%Joshi \textit{et al}. decomposed the image into color, texture and lighting layers, and performed PCA decomposition for each layer.
They performed global corrections of non-facial regions (such as deblurring, color and exposure corrections) using mean and basis vectors generated using PCA decomposition (of every image layer) to derive priors for MAP estimation.
They also performed local corrections of face regions (hallucination for sharpening; or inpainting for exposure correction), by transferring desired properties from HQ images in the gradient domain, using the Poisson equation.

The major drawback of this algorithm is its simplistic model which can address only frontal images with little expression variations and large non-facial regions. We wish to focus on a more high-quality enhancement of \textit{facial} regions, and handle a variety of \textit{subtle} expression variations.

Following this, Loke \textit{et al}.~\cite{loke2013face} suggested to  super-resolve very LR facial images by selecting a set of the most similar HR same-identity training images, in the sense of pose and expression. A similarity measure, based on pose estimation and an expression descriptor, relying on shape and texture, was used for selection. After aligning the selected images using triangulation and affine warping, patches of them were used to hallucinate the face using a MRF model, based on color and edge constraints and a smoothness term.

Drawbacks of this work include the selection process, based on a rough match of some facial regions to the query; we wish to handle more \textit{subtle} expression variations. Replacing LR patches with HR ones results in noticeable artifacts, seams and change of color, since this patch-based method does not account for the human observer's sensitivity to certain facial regions and their expressions. Other drawbacks are using a very large HR dataset (thousands of images), their small size, and the manual labeling of feature points in the LR image. %As we have mentioned, we aim to handle more high-resolution images.
\subsection{Insights}
Previous works and early experiments point our important insights regarding facial images of a specific individual.
\begin{itemize}
	\item The \textit{non-rigid} behavior of faces and facial features under expression variation requires \textit{non-rigid} registration, rather than affine. Most non-rigid methods do not use landmarks but pixels' intensities directly, since they need denser image information, and geometric landmarks are not invariant under non-rigid transformations~\cite{cachier1999fast}; e.g., locations of facial interest points under expression variations.% (see Fig. \ref{fig::zhu_errors}).
	\item As mentioned before, Joshi \textit{et al}. observed that the space spanned by same-identity facial images, depicting a limited range of expressions, is significantly smaller than that spanned by multiple-identity images. Using \textit{generic} faces as priors, on the other hand, introduces artifacts and possible changes in identity and expression. %But using PCA for image reconstruction based on multiple same-identity images might result in a change of facial expression.
	\item A change in identity or facial expression is visually very disturbing to a human observer. Therefore, only the most suitable examples, in the sense of shape, expression, gaze etc., should be used for reconstruction (This can also be deduced from~\cite{jia2008generalized},~\cite{joshi2010personal} and~\cite{loke2013face}).
	\item As Capel and Zisserman have observed, better learning is obtained when considering different facial regions, rather than \textit{the whole face}. Loke \textit{et al}.'s results demonstrate potential difficulties using a \textit{patch-based} method, which does not %doesn't
take into account human observer sensitivity to certain facial regions and their expressions. Capel and Zisserman also noticed that better representation is needed when handling high-detail facial regions that attract human attention and convey facial expression, such as eyes, compared to smooth regions, such as cheeks.
	\item Decomposing the face into facial regions increases the versatility in generating a variety of possible expressions, while decreasing the number of samples required. Since a certain "eye mode" (gaze, shape and closure) can be "accompanied" by many mouth expressions, this decomposition allows to construct and search datasets of small facial regions, rather than large whole-face images, saving both memory and computation time.
\end{itemize}
\subsection{The Proposed Method}
In our work we use personal priors to enhance the quality of facial images of a particular person. We obtain new data-driven facial features spaces, based on only tens of high-quality, same-identity, same-pose example images, differing in facial expression; and define a new affinity measure to match them to given poor-quality images.

%We first extract different key facial features of a certain head pose.
For each key facial feature (eye and mouth) and for different head poses, we construct a high-quality, identity-specific affinity space, representing various different "principal modes" of the specific feature, such as different eye gaze, closure and shape, or different mouth expressions (Fig. \ref{fig::space_illustration}). This is done using a newly defined affinity measure for image matching under non-rigid variations, which derives from the distance between images, in the sense of the diffusion-based Demon transformation~\cite{thirion1995fast} required to register them.

This measure corresponds to the "visual validity" of images interpolated during the diffusion process: how natural, real-world they appear to a human observer. Fluid registration can also interpolate real-world looking images, that can expand the affinity space. Demon registration also provides a useful tool for fine registration of non-rigid facial features.

Given these identity-specific affinity spaces we enhance low-quality, same-identity facial images; specifically, dark cellular phone images degraded by unknown %shot
noise, resolution reduction, slight motion blur and color change. The measure's robustness to quality degradation enables to accurately match input facial features to the most similar example from the corresponding affinity space. Input facial regions are then replaced by the most suitable, Demon-registered, high-quality examples to obtain a high-quality %, same-pose, same-expression
facial image (Fig. \ref{fig::flow_algorithm}).

\section{Demon diffusion-based affinity space}
\subsection{Demon Diffusion-Based Fluid Registration}
The Demon registration, first introduced by Thirion~\cite{thirion1995fast},~‎\cite{thirion1998image}, describes the gradual diffusion process of an object, represented by a deformable grid, into another object, represented by a semi-permeable membrane, through its boundaries by the action of Demon effectors.

Thirion showed the translation of this concept into a simple gradient-based displacement field $\vec{u}$ from the \textit{moving image} $m$ to the \textit{static image} $s$.‎ The improvements suggested by Wang \textit{et al}.~‎\cite{wang2005validation} and Cachier \textit{et al}.~‎\cite{cachier1999fast} yield the following:
\begin{equation}
\vec{u} = (m-s) \times \left (\frac {\vec\nabla s}{|\vec\nabla s|^2 + \alpha^2 (s-m)^2}+\frac {\vec\nabla m}{|\vec\nabla m|^2 + \alpha^2 (s-m)^2}\right)\label{Demon},
\end{equation}
where $\vec\nabla$ denotes image gradient, and $\alpha$ is a normalization factor accounting for adaptive force strength adjustment.

This registration method was so far usually used for medical image registration, such as the work of Kroon and Slump~\cite{kroon2009mri}, whose implementation we use. A detailed explanation is given in Appendix \ref{Demon Registration}.
\subsection{Demon Diffusion-Based Affinity Measure} \label{measure}
\begin{figure}[!h]
	\centering
	\footnotesize
	\begin{tabular}{
			>{\centering\arraybackslash}m{3.5in}}
		\textcolor{green}{\textit{Real-world looking}} change of eye gaze, $D_T=130.56$\\
		\includegraphics[width=3.5in]{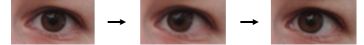}
		\label{fig::valid1} \\
		\textcolor{green}{\textit{Real-world looking}} change of eye shape, $D_T=253.3$ \\
		\includegraphics[width=3.5in]{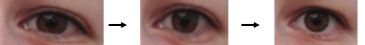}
		\label{fig::valid3} \\
		\textcolor{red}{\textit{Distorted}} change of eye gaze, $D_T=486.76$ \\
		\includegraphics[width=3.4in]{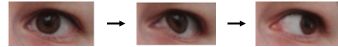}
		\label{fig::non_valid1} \\
%		Change of eye gaze, identity \#2, $D_T=252.3$\\
%		\includegraphics[width=3.5in]{valid2}
%		\label{fig::valid2} \\
%		Change of mouth shape, identity \#2, $D_T=646.23$\\
%		\includegraphics[width=3.5in]{valid5}
%		\label{fig::valid5}	
	\end{tabular}
	\vspace{-0.2cm}	
	\begin{tabular}{
			>{\centering\arraybackslash}m{1in}
			>{\centering\arraybackslash}m{1.2in}
			>{\centering\arraybackslash}m{0.9in}}
		source image & interpolated image & target image	\\
	\end{tabular}
	\caption{Examples of deformation paths between high-quality eye images for identity \#1, and their Demon measures. Compare the \textbf{low} Demon measure distances for the \textcolor{green}{\textit{visually valid}} paths to the \textbf{higher} distance for the \textcolor{red}{\textit{visually non-valid}} path.}
	\label{fig::eyes_visuality}
\end{figure}

We construct an affinity measure to characterize the sequence of intermediate images generated during Demon diffusion (the "deformation path"), by its "visual validity": how natural, real-world and undistorted the path appears to a human observer (Fig. \ref{fig::eyes_visuality}%,\ref{fig::mouths_visuality})
). We will show the resulting high correspondence between the measure and visual validity.

%% REMOVED FIG 3, (GG)
\off{
\begin{figure}[!h]
	\centering
	\footnotesize
	\begin{tabular}{
			>{\centering\arraybackslash}m{3.5in}}
		\textcolor{green}{\textit{Real-world looking}} change of mouth shape, $D_T=1133.3$ \\
		\includegraphics[width=3.5in]{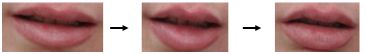}
		\label{fig::valid4} \\
%		\vspace{0.2cm}
		%		Drastic change of eye shape and closure \\
		%		\includegraphics[width=3.5in]{non_valid2}
		%		\label{fig::non_valid2} \\
		%		Drastic change of eye shape and closure \\
		%		\includegraphics[width=3.5in]{non_valid3}
		%		\label{fig::non_valid3} \\
		\textcolor{red}{\textit{Distorted}} change of mouth shape, $D_T=1650.1$ \\
		\includegraphics[width=3.4in]{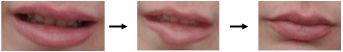}
		\label{fig::non_valid4} 	
		\vspace{0cm}
	\end{tabular}
	\begin{tabular}{
			>{\centering\arraybackslash}m{1in}
			>{\centering\arraybackslash}m{1.3in}
			>{\centering\arraybackslash}m{0.8in}}
		source image & interpolated image & target image	\\
	\end{tabular}
	\caption{Examples of deformation paths between high-quality mouth images for identity \#1, and their Demon distances \GG{$D_T$, Eq. \eqref{Demon_measure}}. Compare the \textbf{low} Demon distance for the \textcolor{green}{\textit{visually valid}} path to the \textbf{higher} distance for the \textcolor{red}{\textit{visually non-valid}} path.}
	\label{fig::mouths_visuality}
\end{figure}
}
To this end, we define a new Demon-based affinity measure (Eq. \eqref{Demon_measure}) of the similarity between images under non-rigid variations. It is derived from the distance between images, in the sense of the Demon transformation required to register them. We use it as an affinity measure and matching criterion of different principal modes of the same facial feature.

The distance measure between two images is proportional to the mean absolute error between the \textit{deformed} image $m$ at a \textit{fixed} time point $T$ in the registration process (taken in our implementation as 200 iterations), and the target image $s$:
%\begin{equation}
%D_T(m,s)=\frac {10^3}{MN}\sum_{i \in [1,M]} \sum_{j \in [1,N]} |m(i,j,t_0)-s(i,j)|\label{Demon_measure}
%\end{equation}
%{\color{red} I SUGGEST TO CHNAGE $t_0$ to $T$.}
%\begin{multline}
%D_T(m,s,\alpha_l)= \\
% \frac {C}{MN}\sum_{i \in [1,M]} \sum_{j \in [1,N]} |m_{HSV,\alpha_l}(i,j,T)-s_{HSV,\alpha_l}(i,j)|\label{Demon_measure}
%\end{multline}
\begin{multline}
D_T(m,s,\alpha_l)=C\|m_{HSV,\alpha_l}(T)-s_{HSV,\alpha_l} \|_{L_1}\label{Demon_measure}
\end{multline}
Where $\alpha_l$ indicates feature-dependent HSV color space channel selection: hue channel for mouths, value channel for eyes.
Intuitively, it is a measure of the distance "left to go" from $m$ to $s$; taking into consideration not only their na\"{\i}ve” pixel-to-pixel similarity, but also Demon's \textit{ability} to successfully deform one into another \textit{in a given time} (as opposed to Cachier's minimization criterion, see Appendix \ref{Demon Registration}). It also relates to local structure and shape (as opposed to histogram distances or EMD, relating to \textit{global}, non-spatial color information).

Given all these characteristics, it better reflects human visual judgment of visual validity, as perceived by a human observer, which can be roughly classified into two categories:
\begin{enumerate}
	%	\item \textit{A visually valid, yet uninteresting, deformation path}. That is, the source and target images are too similar, so that the interpolated images along the deformation path do appear as real-world images, yet do not reflect any interesting deformation process, as seen in Fig. \ref{fig::uninteresting}. This case corresponds to low values of the quantitative distance.
	\item \textit{A visually valid deformation path}. Interpolated images along the deformation path appear as real-world images, describing intermediate phases in the gradual deformation process between images. This case corresponds to lower values of the quantitative distance.
	\item \textit{A visually non-valid deformation path}. Interpolated images along the deformation path appear distorted and cannot be considered as real-world facial features. This case corresponds to higher values of the  distance.
\end{enumerate}
Fig. \ref{fig::eyes_visuality} shows examples of deformation paths between high-quality eye images, and their Demon measures. It can be seen that the \textcolor{green}{\textit{visually valid}}, real-world looking paths, depicting interesting, moderate variations, such as changes in eye gaze or shape, correspond to lower Demon distances; whereas the \textcolor{red}{\textit{visually non-valid}}, distorted looking deformation path corresponds to a higher Demon distance.
%Fig. \ref{fig::mouths_visuality} shows similar correspondences for mouth images.

\begin{figure}[!t]
	\captionsetup{justification=centering}
	\centering
	\includegraphics[width=3.5in]{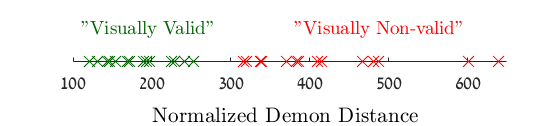} %former: visual_demon_graph, narrow option: visual_demon_graph3
	\caption{Correspondence between visual validity and the (normalized) Demon distance (for eyes of identity \#1). Compare the \textbf{lower} Demon distances, corresponding to the \textcolor{green}{\textit{visually valid}} interpolated images, to the \textbf{higher} Demon distances, corresponding to the \textcolor{red}{\textit{visually non-valid}} images.}
	\label{fig::visual_demon_graph}
\end{figure}

Fig. \ref{fig::visual_demon_graph} shows the correspondence between visual validity and the Demon distance for eye images of identity \#1 (examples of which appear in Fig. \ref{fig::eyes_visuality}). Visual validities of different deformation paths were determined using the concept described above. It can be seen that \textbf{lower values of the Demon distance correspond to visually valid deformation paths, and vice versa}; thus, different visual validity categories can be automatically differentiated using this measure.
Note that we later use a nearest-neighbor scheme to choose the most relevant patch, so no actual threshold of validity should be chosen.

Appendix \ref{synthetic} shows a similar behavior when deforming synthetic images: for moderate variations, deformation succeeds and the measure \textit{moderately} increases with variation. But for more drastic variations, the deformed image becomes too different or distorted; and the measure \textit{drastically} increases.

\begin{figure}
	\centering
	\footnotesize
	\begin{tabular}{
			>{\centering\arraybackslash}m{3.3in}}
		Demon deformation between images of similar illumination \\
		\includegraphics[width=3.3in]{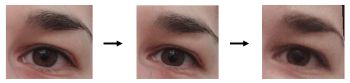}
		%		\label{fig::same_colors}
		\vspace{-0.4cm}
	\end{tabular}	
	\begin{tabular}{
			>{\centering\arraybackslash}m{1in}
			>{\centering\arraybackslash}m{1in}
			>{\centering\arraybackslash}m{1in}}
		HQ source image & \textcolor{green}{\textit{real-world looking}} interpolated image & illumination-adjusted LQ target image
		\vspace{0.4cm}
	\end{tabular}
	\begin{tabular}{
			>{\centering\arraybackslash}m{3.3in}}
		Demon deformation between images of different illumination \\
		\includegraphics[width=3.3in]{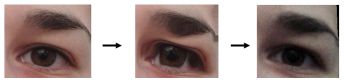}
		%		\label{fig::diff_colors}
	\end{tabular}	
	\begin{tabular}{
			>{\centering\arraybackslash}m{1in}
			>{\centering\arraybackslash}m{1in}
			>{\centering\arraybackslash}m{1in}}
		HQ source image & \textcolor{red}{\textit{distorted}} interpolated image & LQ target image
	\end{tabular}
	\caption{Effect of illumination adjustment on Demon deformation: the naturally-looking, high-quality image obtained when deforming same-illumination images (top); compared to the distorted image obtained when deforming different-illumination images (bottom).}
	\label{fig::color_effect}
\end{figure}

Illumination consistency between images has much influence on Demon registration distortion (Fig. \ref{fig::color_effect}). Registering images similar in shape and structure, but differing in illumination, results in a distorted interpolated image, compared with the naturally-looking, same-structure, high-quality result obtained when a simple illumination adjustment (histogram equalization) is first used.

The measure cannot be considered as a distance or metric in the mathematical sense, as a triangle inequality cannot be shown.
%However, even though the deformation path is directional (resulting in the measure's asymmetry), its direction usually doesn't affect its visual validity.
%As for now, it seems \textit{nonparametric}; %, except for its dependence in the number of iterations and inner Demon parameters.
%that is, no specific parameters can be tuned to produce a change in eye gaze, for instance; or to \textit{extrapolate} images.

\begin{figure*}[!h]
	\captionsetup{justification=centering}
	\centering
	\includegraphics[width=5in]{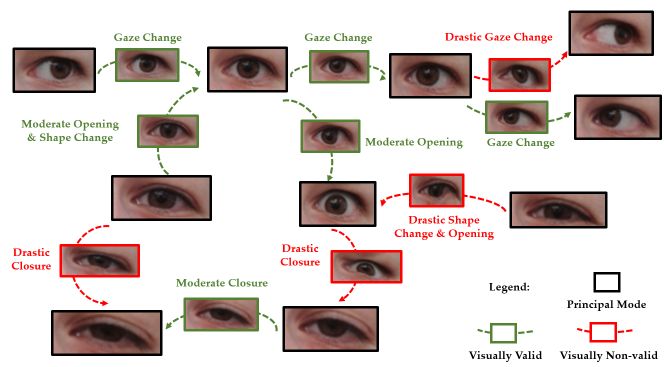}
	\caption{Illustration of an eye affinity space, constructed based on the Demon-based affinity measure. Visually valid deformation paths and interpolated images appear in green, whereas visually non-valid ones appear in red.}
	\label{fig::space_illustration}
\end{figure*}
A visually valid deformation path allows using the interpolated images as new real-world looking images, describing intermediate phases of subtle variations between existing principal modes, thus %potentially
expanding considerably the available dataset.
Finally, \textit{Demon registration enables fine non-rigid registration} of facial features, needed to handle their non-rigid behavior under expression variations.

Concluding the Demon registration and measure important characteristics:
\begin{enumerate}
	\item \textbf{Correspondence between Demon measure and visual validity}: As Demon registration relates to \textit{shape and structure}, the deformed image reflects Demon's ability to handle non-rigid image variations, while preserving real-world appearance. As the Demon measure relates to Demon's ability to bring one image close to another, rather then their original distance, it corresponds to the visual validity of the deformed image resulting from registration. Therefore, \textit{low Demon distances} correspond to \textit{moderate non-rigid variations} between images, with \textit{real-world appearance} of the interpolated images.
	\item \textbf{Robustness to quality degradation}: Demon registration is quite robust to quality degradations, such as noise and resolution reduction, given that illumination is consistent; therefore, when registering HQ images to an LQ image, low distances still correspond to similar structures.
	\item \textbf{Preserving source quality when registering different-quality images}: registering a HQ image to a similar-shape LQ image preserves its high quality, while adjusting to the desired shape, as can be seen in Fig. \ref{fig::color_effect}.
\end{enumerate}
%{\color{red} removed the bold-faces, a little too much..}
Combining these characteristics allows performing a {measure-based Nearest Neighbor search} to match a LQ query image to the {most similar HQ dataset image}, in the sense of Demon registration. The {interpolated image} resulting from registering the HQ match to the LQ query is of quite {high-quality, naturally-looking and of desirable shape}.
%Note, that image registration was done using feature-dependent selection of the HSV channel (Figs. \ref{fig::non_valid}, \ref{fig::valid}).
%Note the use of an \textit{absolute} error-based measure, rather than a \textit{squared} error-based measure, which doesn't well correspond to the visual validity. A possible reasoning could be that MAE measure is practically a size-normalized version of L1-norm, which reduces outliers.
%The measure is not necessarily symmetric, and the triangle inequality does not always hold; therefore, it
% When registering a HQ image to a LQ image,
%% shape and structure adjust to the desired ones, but image quality is quite well preserved, so that
%the resulting image is of desirable shape and structure, as well as quite high quality (see Fig. \ref{fig::color_effect}).
%{\color{red} removed the last conclusion, somewhat repetitive..}

%{\color{red} STOPPED CORRECTING HERE.}
\off{
In conclusion, the three merits of using the Demon registration and the related affinity measure are:
\begin{enumerate}
	\item A newly defined affinity measure for image matching, which corresponds to the visual validity of interpolated images, and is quite robust to quality degradation.
	\item Generating real-world looking visually valid \textit{interpolated} images along the deformation path, that can be used to expand the dataset.
	\item Fine quality-preserving, non-rigid feature registration.
\end{enumerate}
}
\subsection{Demon-Based Facial Features Affinity Spaces}
\begin{figure}[!h]
	\captionsetup{justification=centering}
	\centering
	%	\subfloat[]{\includegraphics[height=2in]{mouth_space1}%
	%		\label{fig::mouth_space1}}
	%	\qquad
	%	\subfloat[]{\includegraphics[height=2in]{mouth_space2}%
	%		\label{fig::mouth_space2}}
	%	
	%	\subfloat[]{\includegraphics[width=5in]{eye_space}%
	\includegraphics[width=3.5in]{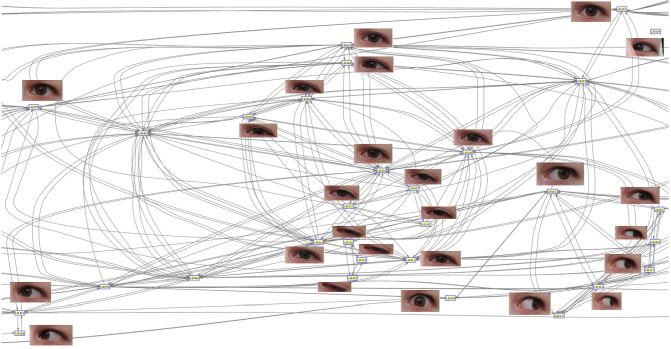}
	%		\label{fig::eye_space}}
	%	\hfil
	\caption{Example of a real affinity space of same-identity, same-pose eyes, automatically constructed using the Demon affinity measure.}
	\label{fig::real_spaces}
\end{figure}

Fig. \ref{fig::space_illustration} illustrates the concept of an affinity space based on the Demon-based affinity measure, with visually valid deformation paths between principal modes.
%Each space describes Demon-based relations between different principal modes:
%describing a variety of facial expressions of the \textit{specific} facial feature.
%visually valid deformation paths connect different principal modes
A geodesic of interpolated, visually valid images, depict intermediate steps in the deformation between images. Visually non-valid deformation paths are not allowed, since their interpolated images cannot be used to generate new visually valid images.
Fig. \ref{fig::real_spaces} shows a real affinity space of same-identity, same-pose eyes, automatically constructed using the Demon affinity measure. Note that a non-frontal eye (uppermost right) wrongly classified as frontal during preprocessing (see Sec. \ref{sec::Preprocessing}) is unconnected to all others. The affinity measure is used as a \textit{matching criterion for choosing} an image from the dataset, which is the \textit{most} similar to a given test image. \textbf{Note, that this search does not require knowledge of the connections between dataset images, or paths' visual validities}. As we have seen, the measure's robustness to image quality degradation allows finding the most suitable match even for a poor-quality query image.
%Note that neither the illustrated nor the real affinity spaces display the many uninteresting paths.
\begin{figure*}[!h]
	\tabcolsep0.3mm
	\begin{tabular}{ccc ccc ccc} 	
		\includegraphics[width=0.11\textwidth]{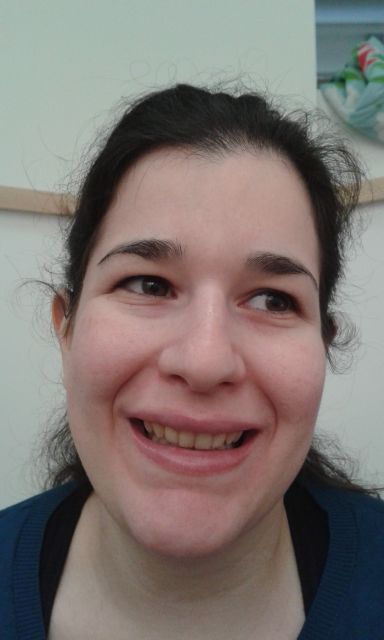} &
		\includegraphics[width=0.11\textwidth]{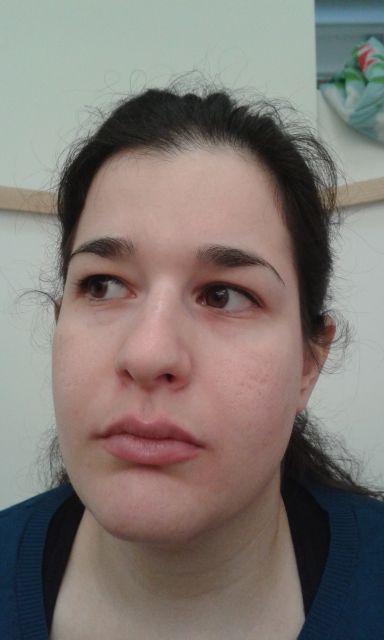} &	
		\includegraphics[width=0.11\textwidth]{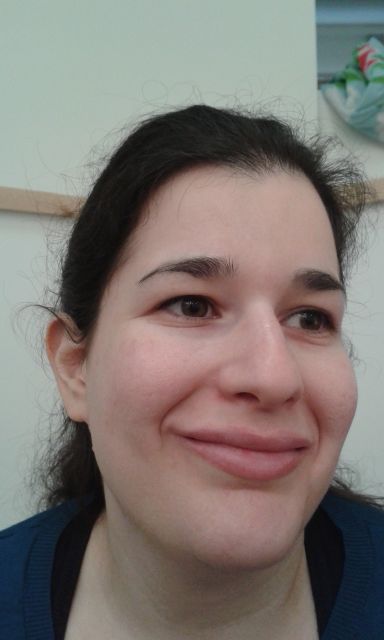} &
		\includegraphics[width=0.11\textwidth]{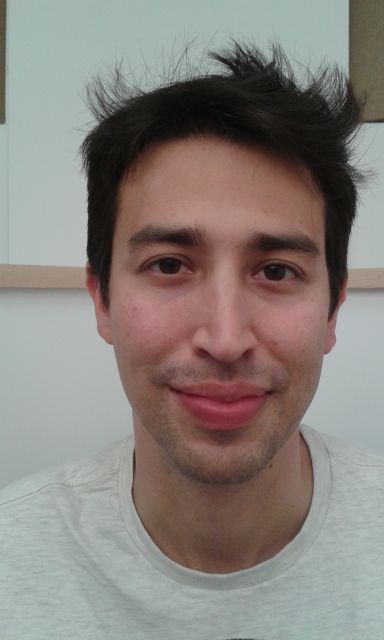} &
		\includegraphics[width=0.11\textwidth]{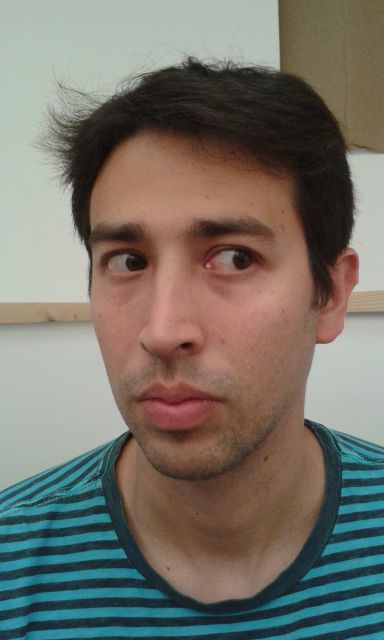} &
		\includegraphics[width=0.11\textwidth]{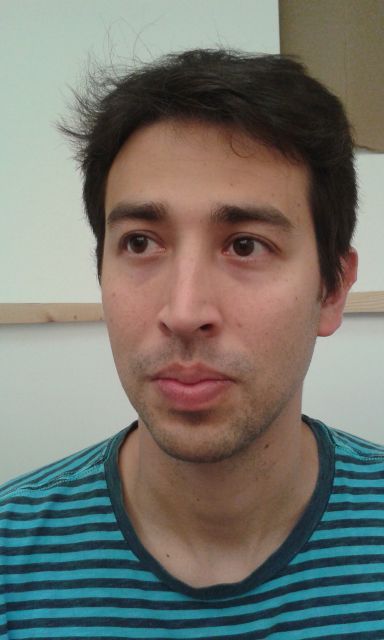} &	
		\includegraphics[width=0.11\textwidth]{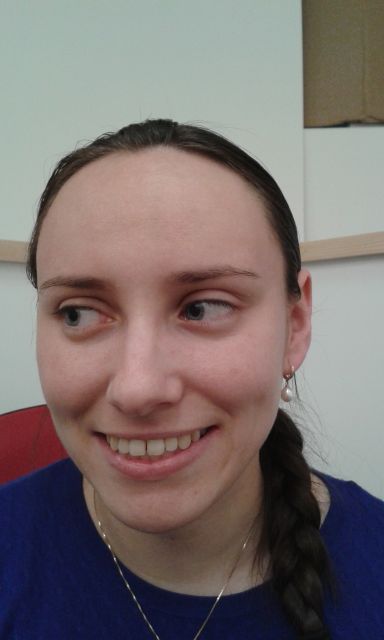} &
		\includegraphics[width=0.11\textwidth]{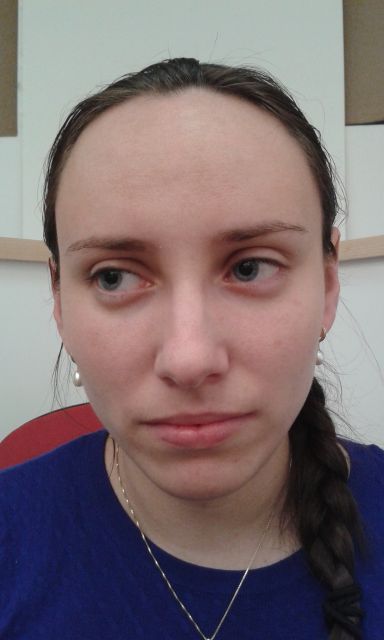} &
		\includegraphics[width=0.11\textwidth]{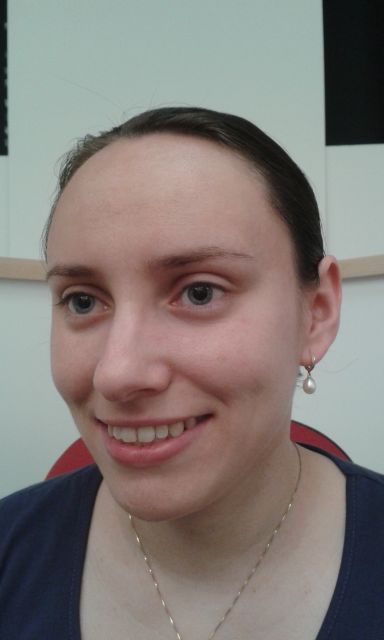}
	\end{tabular}
	\caption{Examples of our personal priors image set, which includes 7 sets of different identities and poses; each consists of 20-30 same-identity, same-pose, multiple-expression high-quality cellular images.}
	\label{fig::imageSet}
\end{figure*}
\begin{figure*}[!h]
	\captionsetup{justification=centering}
	\centering
	\includegraphics[width=6.5in]{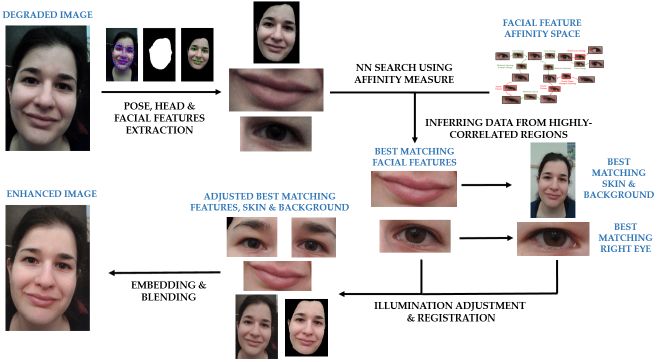}
	\caption{Algorithm's flowchart: Facial image quality enhancement using the Demon-Based affinity measure and affinity spaces.}
	\label{fig::flow_algorithm}
\end{figure*}

We use only tens of personal priors images to \textit{automatically} construct 14 data-driven spaces (2 identities with 2 poses each, one identity with 3 poses; for each pose, two spaces are constructed: for the left eye and for the mouth). Each space consists of multiple high-quality, same-identity, same-pose examples of a specific facial feature (about 20-30 principal modes per space). Fig. \ref{fig::imageSet} shows examples of the personal priors image set.
As opposed to previous works, affinity spaces describe many different \textit{subtle} expression variations, such as different eye gaze, closure and shape, or different mouth closure, shape and expression.
% Therefore, when constructing the space, a newly-arrived example which is too different from all existing principal modes, that is whose deformation paths to the existing principal modes are all visually non-valid, will be stored as a new (meanwhile) \textit{unconnected} principal mode.
%On the other hand, images very similar to an existing principal mode may be stored as sample points \textit{related} to that mode, but not as new principal modes. The use of color properties to improve the construction of the mouth space will be later addressed (see Sec. \ref{mouth}).

In the future, it might be possible to use the subset of the most similar principal modes and the visually valid interpolated images between them as priors or constrains for other frameworks of image restoration. Another option might be to first get a rough notion of the relevant subset, using some kind of an initial "projection" of the degraded test image onto the space, thus making the selection process more time-efficient.

\section{Facial image quality enhancement}
We now show the use of personal priors and the Demon concept for a semantically-aware enhancement of facial images. Fig. \ref{fig::flow_algorithm} illustrates the proposed method.
\begin{algorithm}[!h]
	\KwData{Degraded facial image (blind degradation model), HQ personal priors.}
	\KwResult{Enhanced facial image.}
	Extract facial features, select relevant HQ patch affinity spaces according to identity and pose, Sec. \ref{sec::Preprocessing}\;
%    {\color{red} Correct color of entire image according to..??} \;
	For each semantic patch (eye, mouth): select the most similar HQ patch in the space, using illumination adjustment and Demon measure, Sec. \ref{sec::NN}\;
	Infer data for other facial regions from highly-correlated regions, Sec. \ref{sec::infer}\;
	Embed high-quality image details, using registration, color-correction and blending, Sec. \ref{sec::embed}\;
\caption{Facial Image Quality Enhancement.}
	%{\color{red} The description was too short, very hard to understand what is done, even at a very high level, added a little, correct my remarks where necessary..}}
\end{algorithm}

The details of the algorithm are as follows:
\subsection{Preprocessing: Facial Features Extraction} \label{sec::Preprocessing}
We use Zhu and Ramanan's algorithm~\cite{zhu2012face} to detect the facial contour, whose convex hull is used as input to the image matting algorithm of Levin \textit{et al}.~\cite{levin2008closed}. Thresholding and erosion of the resulting mask (similar to the preprocessing in~\cite{joshi2010personal}) result in a head image, which will be later used for skin texture enhancement. We also use Zhu and Ramanan's pose estimation, to later search the suitable (same pose-sign) affinity spaces. For a more accurate facial landmarks localization we prefer using the algorithm of Asthana \textit{et al}.~\cite{asthana2014incremental}, with the head image as the initial face detection. \textbf{Note, that HQ images are processed similarly for pose estimation, features and head extraction to construct the HQ spaces.}
%{\color{red} This part has many important stages and needs at list a basic illustration, mask of face, facial landmarks etc. To save space maybe this could fit in the first stage of Fig. 8 of the algorithm flowchart (similar to Fig. 1 bottom left).}

%{\color{red} Somewhere here, or in the previous section we should mention in a sentence or two how the HQ images are processed (sort of "training data"). 
%I assume essentially many of the computer-vision algorithms and other processing to extract the eyes/mouth are the same as in the main algorithm.
%}
\subsection{NN Search using Affinity Measure} \label{sec::NN}
%{\color{red} Where is the color being corrected and how? to search for NN you already need it to be with a similar color scheme, right?}
Nearest Neighbor searches through suitable (same pose-sign) affinity spaces are conducted to find the best matching high-quality examples in the dataset. Throughout the search, \textbf{illumination adjustment} (using histogram equalization) is performed prior to distance calculation.
\subsection{Inferring Data from Highly-correlated Regions} \label{sec::infer}
For facial image enhancement, structure and context correlations between semantically meaningful face regions, used as building blocks, are used to infer the suitable data.
To select the proper \textit{right} eye, based on the \textit{left} eyes space, we make the reasonable assumption of gaze, closure and shape consistency between both eyes (ignoring cross-eye and winking). This allows to extract the suitable right eye from \textit{the same high-quality image from which the left eye example was taken}. Thus, avoiding the need to construct a right eyes space.

Another use of facial semantic structural constrains regards head structure and skin texture. In general, the shape of middle-low facial regions (cheeks, chin, facial lower contour and even nose) depends on the mouth expression, but remains unchanged under eye expression variations. Therefore, given the large collection of available high-quality images, it is only reasonable to use \textit{the same high-quality image from which the mouth was taken} to also extract skin information. Therefore, a high-quality background image, head structure and skin texture information are selected according to best matching mouth.
\subsection{Embedding High-quality Image Details} \label{sec::embed}
The \textbf{input image and input head/skin} undergo \textbf{illumination adjustment} to the brighter illumination of the selected high-quality background image and example head/skin image, respectively, using the NRDC algorithm~\cite{hacohen2011non}. Due to the randomized nature of Generalized PatchMatch embedded in NRDC, we choose out of several repetitions the best illuminated background, \textit{in the sense of its NIQE score}~\cite{mittal2013making} (see Sec. \ref{results}).
The \textbf{example head/skin} image undergoes affine \textbf{registration} to best fit the input.
%is first adjusted to the initially enhanced feature using illumination adjustment and fine non-rigid registration, and then is used to replace it.
As to the \textbf{facial features}, input features undergo \textbf{illumination adjustment} to the brighter high-quality illumination. Then, example features undergo \textbf{fine non-rigid registration} to best fit the input feature structure.

Finally, we embed the high-quality skin texture and facial features information into the brighter noisy image, using Burt and Adelson's multi-resolutional \textbf{blending}~\cite{burt1983multiresolution} to produce a seamless, smooth appearance.
\section{Experimental results} \label{results}
%\begin{table}[h]
%	\begin{center}
%		\caption{normalized niqe quality assessment for different methods}
%		\begin{tabular}{|
%			>{\centering\arraybackslash}m{0.3in} |
%			>{\centering\arraybackslash}m{0.3in} | >{\centering\arraybackslash}m{0.9in} | >{\centering\arraybackslash}m{0.6in} | >{\centering\arraybackslash}m{0.5in} |}
%			\hline
%			\textbf{    } & \textbf{Our method} & \textbf{ Prior-based brightened image} & \textbf{Noisy input image} & \textbf{BM3D Denoising}
%%			\parbox{0pt}{\rule{0pt}{0ex+\baselineskip}}\\
%			\\[0ex] \hline
%			Fig.\ref{fig::denoising_142152} & \textbf{1.0187} & 1.1058 & 1.2032 & 1.6097 \\[0ex] \hline
%			Fig.\ref{fig::denoising_141748} & \textbf{1.1274} & 1.2311 & 1.2442 & 1.7672 \\[0ex] \hline
%			Fig.\ref{fig::denoising_152347} & \textbf{1.1858} & 1.2203 & 1.4501
%			& 1.9304 \\[0ex] \hline	
%%			Fig.\ref{fig::denoising_110038} & \textbf{1.2724} & 1.4145 & 1.4757 & 2.0323 \\[0ex] \hline
%			Fig.\ref{fig::denoising_114610} & \textbf{1.0875} & 1.1189 & 1.3085 & 1.6573 \\[0ex] \hline
%%			Fig.\ref{fig::denoising_141832} & \textbf{1.14} & 1.2459 & 1.3007 & 1.8927 \\[0ex] \hline
%			Fig.\ref{fig::denoising_114536} & \textbf{1.2209} & 1.3338 & 1.4289 & 2.0586 \\[0ex] \hline
%			Fig.\ref{fig::denoising_102623} & \textbf{1.2658} & 1.312 & 1.6402 & 2.025 \\[0ex] \hline
%		\end{tabular}	
%		\label{tabular:NIQE}
%	\end{center}
%\end{table}
\begin{figure}
	\captionsetup{justification=centering}
	\centering
	\qquad
	\subfloat[Low quality input image,\hspace{\textwidth}NIQE score=1.2032]{\includegraphics[height=2.2in]{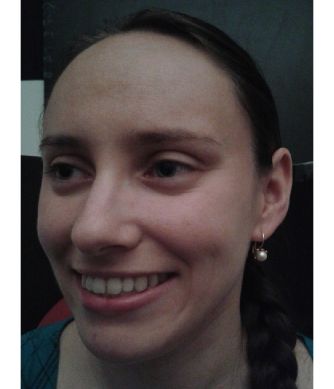}}
	\subfloat[Prior-based brightened input image,\hspace{\textwidth}NIQE score=1.1058]{\includegraphics[height=2.2in]{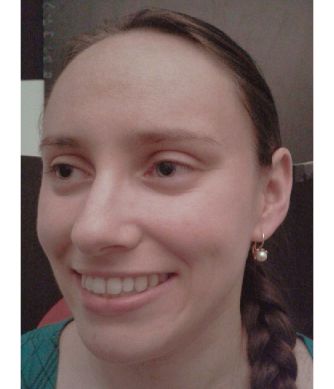}}
	\qquad
	\subfloat[BM3D Denoising of brightened image, estimated noise std=10,\hspace{\textwidth}NIQE score=1.6097]{\includegraphics[height=2.2in]{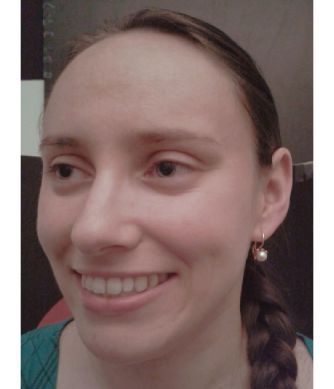}}
	\subfloat[Proposed method,\hspace{\textwidth}NIQE score=1.0187]{\includegraphics[height=2.2in]{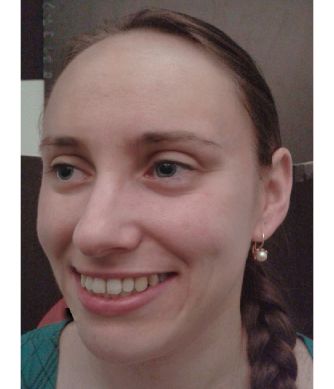}}
	\qquad
	\subfloat[HQ exmaple for mouth \& head info. and background illumination]{\includegraphics[height=1.35in]{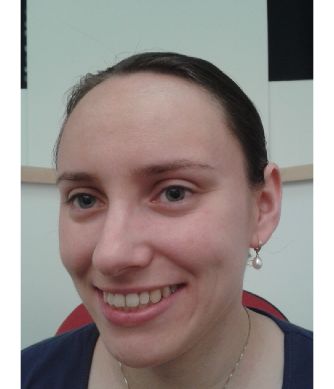}}
	\subfloat[HQ exmaple for eyes info.] {\includegraphics[height=1.35in]{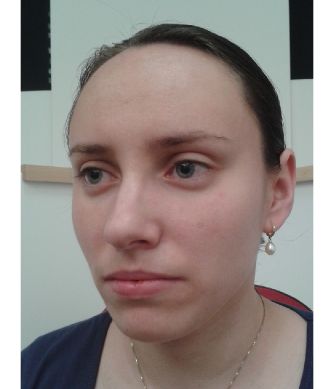}}
	\subfloat[Difference image: Brightened input to our result]{\includegraphics[height=1.35in]{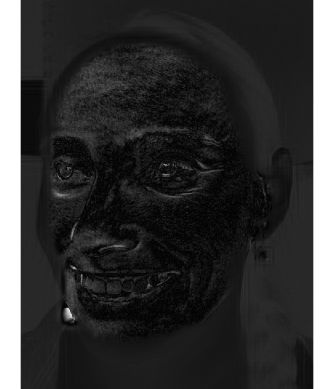}}
	\vspace{0.4cm}	
	\tabcolsep0.3mm
	\begin{tabular}{cccc} 	
		\rotatebox{90}{\hspace{0.2cm}{\small left eye}} &	
		\includegraphics[width=0.13\textwidth]{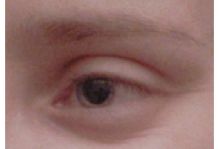} &	
		\includegraphics[width=0.13\textwidth]{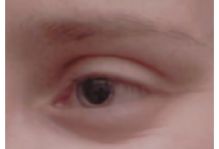} &
		\includegraphics[width=0.13\textwidth]{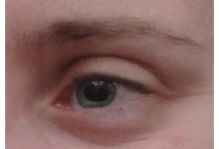}  \\[1mm]
		\rotatebox{90}{\hspace{0.5cm}{\small right eye}} &		
		\includegraphics[width=0.13\textwidth]{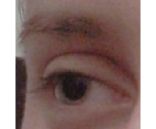} &	
		\includegraphics[width=0.13\textwidth]{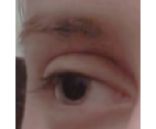} &
		\includegraphics[width=0.13\textwidth]{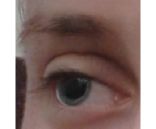}  \\[1mm]	
		\rotatebox{90}{\hspace{0.2cm}{\small mouth}} &	
		\includegraphics[width=0.13\textwidth]{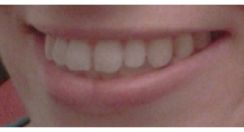} &	
		\includegraphics[width=0.13\textwidth]{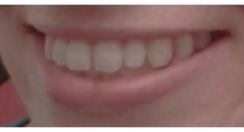} &
		\includegraphics[width=0.13\textwidth]{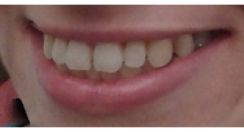}  \\
		& {\small Brightened input} & {\small BM3D Denoising} & {\small Proposed method}
	\end{tabular}\\[1mm]	
	\caption{Denoising and quality enhancement example.}
	\label{fig::denoising_142152}
\end{figure}

\begin{figure}[!h]
	\captionsetup{justification=centering}
	\centering
	\qquad
	\subfloat[Low quality input image,\hspace{\textwidth}NIQE score=1.2442]{\includegraphics[height=2.2in]{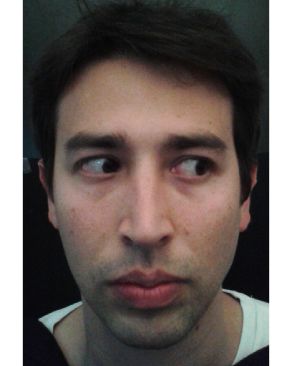}}
	\subfloat[Prior-based brightened input image, NIQE score=1.2311]{\includegraphics[height=2.2in]{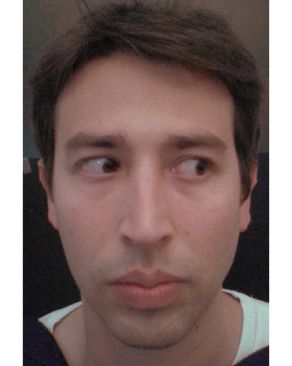}}
	\qquad
	\subfloat[BM3D Denoising of brightened image, estimated noise std=10,\hspace{\textwidth}NIQE score=1.7672]{\includegraphics[height=2.2in]{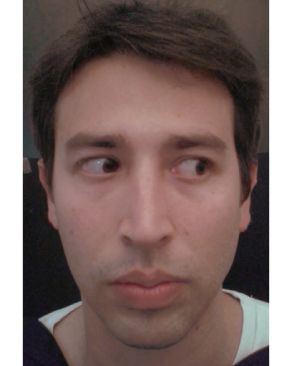}}
	\subfloat[Proposed method,\hspace{\textwidth}NIQE score=1.1274]{\includegraphics[height=2.2in]{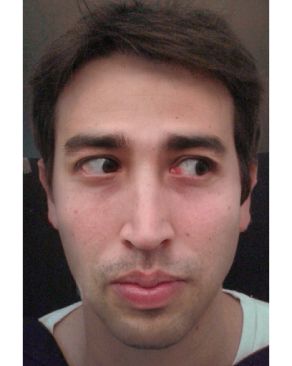}}
	\qquad
	\caption{Denoising and quality enhancement example.}
	\label{fig::denoising_141748}
\end{figure}

\begin{figure}[!h]
	\captionsetup{justification=centering}
	\centering
	\includegraphics[height=0.38\textwidth]{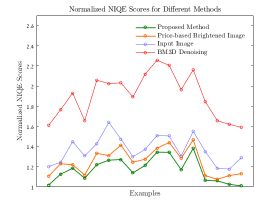}
	\caption{Normalized NIQE scores for different methods for 17 examples. \textbf{The closer the normalized score is to 1 - the better the quality}. Therefore it can be seen that \textcolor{green}{our method} outperforms \textcolor{orange}{the prior-based brightening method}; which outperforms \textcolor{blue}{the input image}; which outperforms \textcolor{red}{BM3D}.}
	\label{fig::NIQE_graph_17}
\end{figure}

%\begin{figure}
%	\captionsetup{justification=centering}
%	\centering
%%	\qquad
%	\subfloat[Low quality input image]{\includegraphics[height=2.4in]{110038_dark}}
%	\subfloat[Prior-based brightened input image]{\includegraphics[height=2.4in]{110038_bright_new}}
%	\qquad
%	\subfloat[BM3D Denoising of brightened image, estimated noise std=10]{\includegraphics[height=2.4in]{110038_BM3D_10_new}}
%	\subfloat[Proposed method]{\includegraphics[height=2.4in]{110038_ours_new}}
%	\qquad
%	\tabcolsep0.3mm
%	\begin{tabular}{ccc}
%		\includegraphics[height=0.22\textwidth]{110038_ours_new_diff2} &
%		\includegraphics[height=0.22\textwidth]{110038_no_face_reg} &		
%		\includegraphics[height=0.22\textwidth]{110038_no_blend} \\[1mm]
%		{\small Difference image} & {\small No head reg.} & {\small No blending}
%	\end{tabular}\\[1mm]
%	\caption{Shot noise denoising and quality enhancement example.}
%	\label{fig::denoising_110038}
%\end{figure}

\begin{figure}[!h]
	\captionsetup{justification=centering}
	\centering
	\subfloat[Low quality input image,\hspace{\textwidth}NIQE score=1.4501]{\includegraphics[height=2.2in]{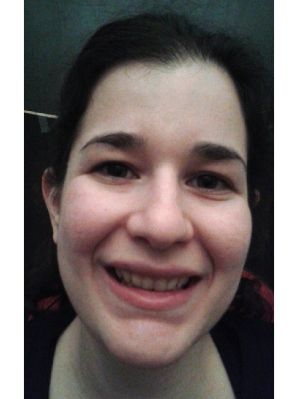}}
	\subfloat[Prior-based brightened input image, NIQE score=1.2203]{\includegraphics[height=2.2in]{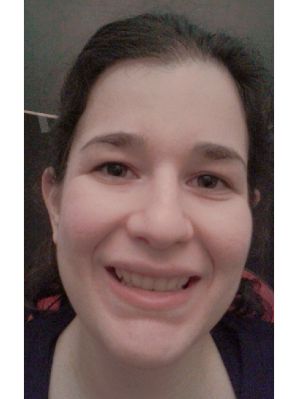}}
	\qquad
	\subfloat[BM3D Denoising of brightened image, estimated noise std=10,\hspace{\textwidth}NIQE score=1.9304]{\includegraphics[height=2.2in]{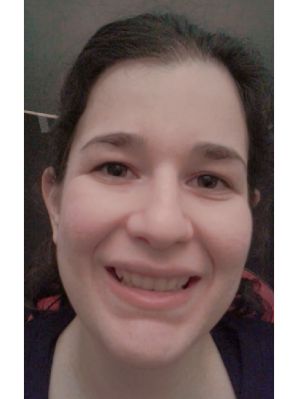}}
	\subfloat[Proposed method,\hspace{\textwidth}NIQE score=1.1858]{\includegraphics[height=2.2in]{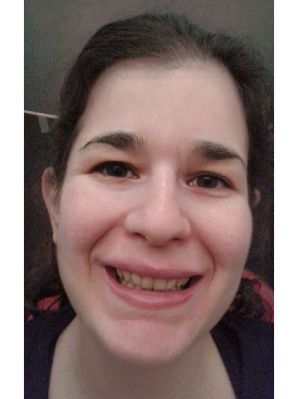}}
	\vspace{0.4cm}
	\tabcolsep0.3mm	
	\begin{tabular}{ccc}
		\includegraphics[height=0.22\textwidth]{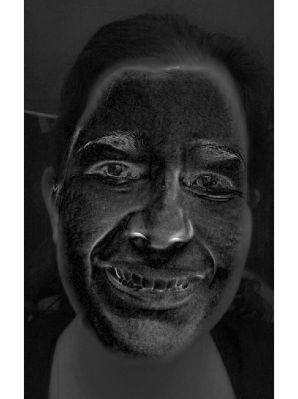} &
		\includegraphics[height=0.22\textwidth]{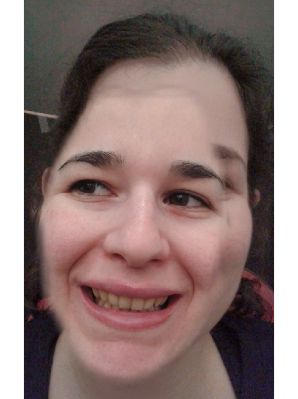} &		
		\includegraphics[height=0.22\textwidth]{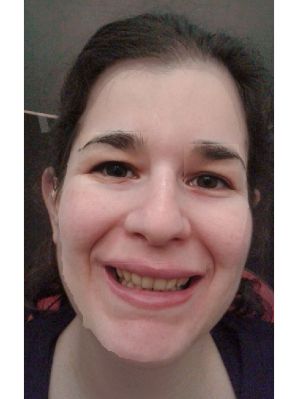} \\[1mm]
		{\small Difference image} & {\small No head reg.} & {\small No blending}
	\end{tabular}\\[1mm]
	\caption{Denoising and quality enhancement example.}
	\label{fig::denoising_152347}
\end{figure}

\begin{figure}[!h]
	\captionsetup{justification=centering}
	\centering
	\subfloat[Low quality input image,\hspace{\textwidth}NIQE score=1.3085]{\includegraphics[height=2in]{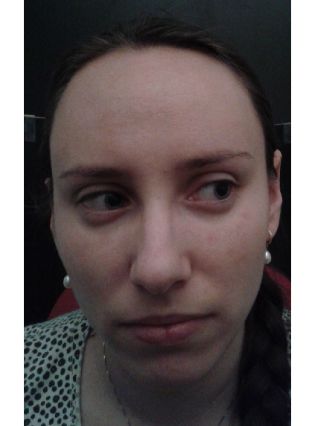}}
	\subfloat[Prior-based brightened input image, NIQE score=1.1189]{\includegraphics[height=2in]{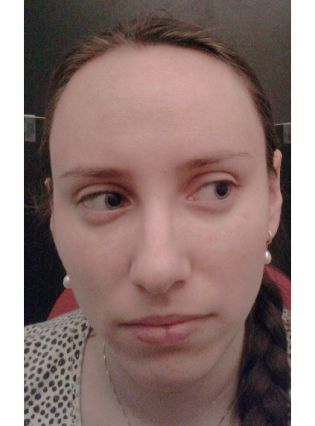}}
	\\
	\subfloat[BM3D Denoising of brightened image, estimated noise std=10, NIQE score=1.6573]{\includegraphics[height=2in]{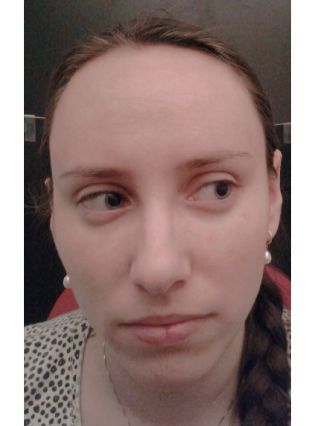}}
	\subfloat[Proposed method,\hspace{\textwidth}NIQE score=1.0875]{\includegraphics[height=2in]{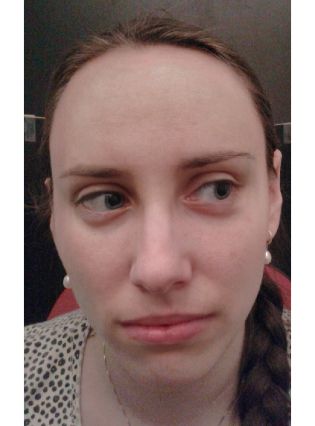}}
	\qquad	
	\caption{Denoising and quality enhancement example.}
	\label{fig::denoising_114610}
\end{figure}

\begin{figure}[!h]
	\tabcolsep0.3mm
	\begin{tabular}{ccc}
		\rotatebox{90}{\hspace{0cm}{\small right choice}} &	
		\includegraphics[height=0.08\textwidth]{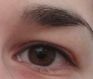} &
		\includegraphics[height=0.08\textwidth]{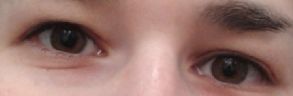} 		\\[1mm]	
		\rotatebox{90}{\hspace{-0.1cm}{\small moderate error}} &
		\includegraphics[height=0.08\textwidth]{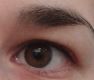} &
		\includegraphics[height=0.08\textwidth]{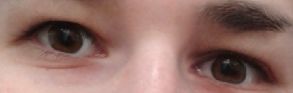} \\[1mm]
%		\rotatebox{90}{\hspace{-0.1cm}{\small drastic error}} &
%		\includegraphics[height=0.08\textwidth]{eye4_for110038} &
%		\includegraphics[height=0.08\textwidth]{110038_witheye4} \\[1mm]
		\rotatebox{90}{\hspace{0cm}{\small drastic error}} &	
		\includegraphics[height=0.08\textwidth]{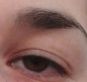} &
		\includegraphics[height=0.08\textwidth]{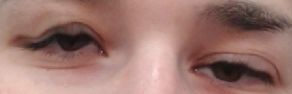} \\[1mm]
%		\rotatebox{90}{\hspace{0cm}{\small drastic error}} &	
%		\includegraphics[height=0.08\textwidth]{eye13_for110038} &
%		\includegraphics[height=0.08\textwidth]{110038_witheye13} \\[1mm]
		\vspace{0.2cm}
		& {\small example eye} & {\small resulting eyes}

	\end{tabular}\\[1mm]
		\tabcolsep0.3mm
		\begin{tabular}{ccc}
			\rotatebox{90}{\hspace{0.6cm}{\small right choice}} &	
			\includegraphics[height=0.13\textwidth]{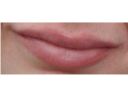}%{rightmouth18_for110038} &
			\hspace{0.2cm}
			\includegraphics[height=0.13\textwidth]{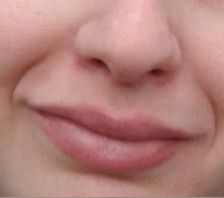} \\[1mm]	
			\rotatebox{90}{\hspace{0.4cm}{\small moderate error}} &
			\includegraphics[height=0.13\textwidth]{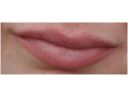}%{mouth17_for110038} &
			\hspace{0.2cm}
			\includegraphics[height=0.13\textwidth]{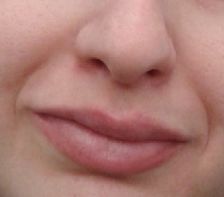} \\[1mm]
%			\rotatebox{90}{\hspace{0.5cm}{\small drastic error}} &
%			\includegraphics[height=0.09\textwidth]{19}%{mouth19_for110038} &
%			\hspace{0.2cm}
%			\includegraphics[height=0.09\textwidth]{110038_withmouth19} \\[1mm]
%			\rotatebox{90}{\hspace{0.6cm}{\small drastic error}} &	
%			\includegraphics[height=0.13\textwidth]{1}%{mouth1_for110038} &
%			\hspace{0.2cm}
%			\includegraphics[height=0.13\textwidth]{110038_withmouth1} \\[1mm]
			\rotatebox{90}{\hspace{0.6cm}{\small drastic error}} &	
			\includegraphics[height=0.13\textwidth]{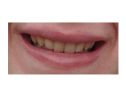}%{mouth29_for110038} &
			\hspace{0.2cm}
			\includegraphics[height=0.13\textwidth]{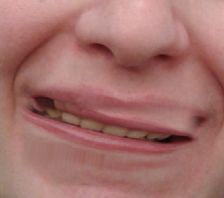}
			\vspace{-0.2cm}
		\end{tabular}\\[1mm]
		\begin{tabular}{
				>{\centering\arraybackslash}m{0.1in}
				>{\centering\arraybackslash}m{1.2in}
				>{\centering\arraybackslash}m{1.2in}}
			& \small example mouth & \small resulting mouth	\\
		\end{tabular}
	\caption{The Demon measure \textbf{allows accurate selection} of example facial features. \textbf{But what if the selection process resulted in errors?} \hspace{\textwidth}Demon registration's \textbf{robustness to moderate non-rigid variations} allows it to \textbf{overcome moderate selection errors}, such that the resulting interpolated features are of quite desirable shape. However, more drastic errors result in severely distorted features. Examples relate to the example in Fig. \ref{fig::flow_algorithm}.}
	\label{fig::errors_110038}
\end{figure}

%\begin{figure}
%	\captionsetup{justification=centering}
%	\centering
%	\subfloat[Low quality input image]{\includegraphics[height=2in]{141832_dark}}
%	\subfloat[Prior-based brightened input image]{\includegraphics[height=2in]{141832_bright}}
%	\\
%	\subfloat[BM3D Denoising of brightened image, estimated noise std=10]{\includegraphics[height=2in]{141832_BM3D_10}}
%	\subfloat[Proposed method]{\includegraphics[height=2in]{141832_ours}}
%	\qquad
%	\caption{Shot noise denoising and quality enhancement example.}
%	\label{fig::denoising_141832}
%\end{figure}

\begin{figure}[!h]
	\captionsetup{justification=centering}
	\centering				
	\subfloat[Low quality input image,\hspace{\textwidth}NIQE score=1.4289]{\includegraphics[height=2in]{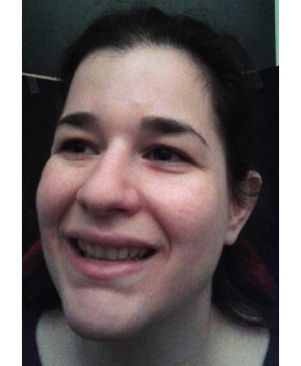}}
	\subfloat[Prior-based brightened input image, NIQE score=1.3338]{\includegraphics[height=2in]{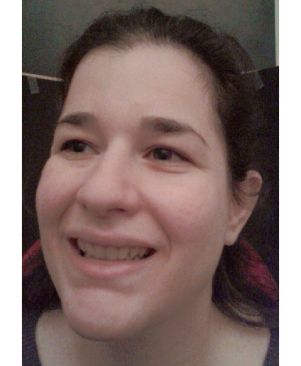}}
	\\
	\subfloat[BM3D Denoising of brightened image, estimated noise std=10, NIQE score=2.0586]{\includegraphics[height=2in]{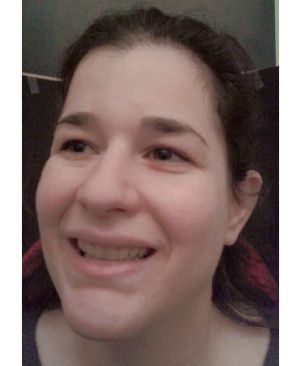}}
	\subfloat[Proposed method,\hspace{\textwidth}NIQE score=1.2209]{\includegraphics[height=2in]{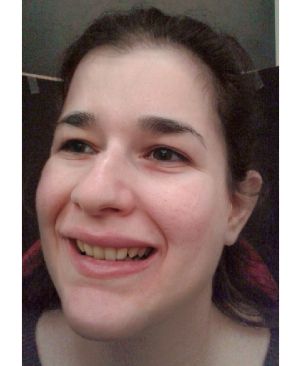}}
	\qquad
	\caption{Denoising and quality enhancement example.}
	\label{fig::denoising_114536}
\end{figure}

\begin{figure}[!h]
	\captionsetup{justification=centering}
	\centering
	\subfloat[Low quality input image,\hspace{\textwidth}NIQE score=1.6402]{\includegraphics[height=2in]{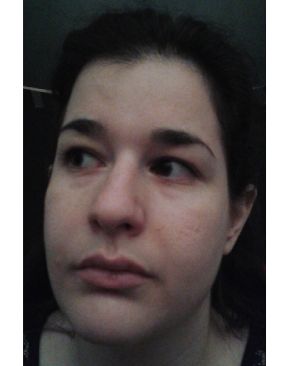}}
	\subfloat[Prior-based brightened input image, NIQE score=1.312]{\includegraphics[height=2in]{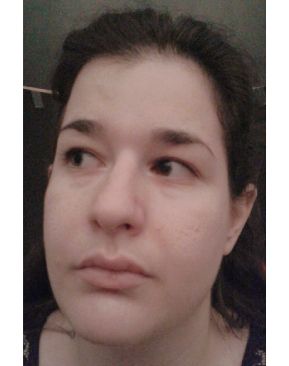}}
	\\
	\subfloat[BM3D Denoising of brightened image, estimated noise std=10, NIQE score=2.025]{\includegraphics[height=2in]{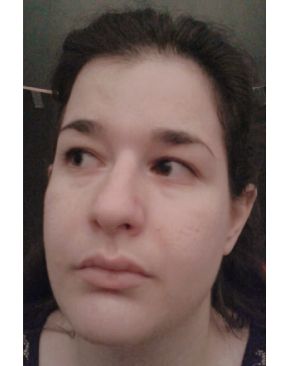}}
	\subfloat[Proposed method,\hspace{\textwidth}NIQE score=1.2658]{\includegraphics[height=2in]{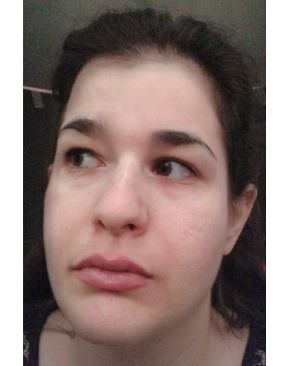}}
	\qquad
	\caption{Denoising and quality enhancement example.}
	\label{fig::denoising_102623}
\end{figure}

We now show experimental results for our prior-based denoising and quality enhancement algorithm.
Input facial images were taken using a SAMSUNG GT-S7580L cellular frontal camera (2560X1536 resolution) in a dark environment. Assuming no known information regarding the camera's specifications and built-in image processing algorithms. This real life scenario of dark environment cellular shooting demonstrates well common flaws of shot noise, post-processed by unknown (possibly nonlinear) filtering, slight motion blur and resolution reduction of unknown parameters. Prior and input images were downsampled by a factor of 2 before processing. Using an unoptimized Matlab code with Matlab/C++ code segments, on a Windows 7 OS, Intel i7-4770 CPU at 3.4 GHz with 16GB RAM, the running time using a single NRDC-NIQE iteration was 3 minutes; running time was 4 minutes when using 5 iterations.

We demonstrate our results for multiple identities, poses and expressions, visually comparing them to the prior-based brightened image using NRDC, and a state-of-the-art BM3D~\cite{dabov2007color} denoising of the brightened image, assuming AGW noise of std=10. Since \textbf{no ground truth images are available}, we use the \textit{no-reference} blind-model image quality assessment score NIQE~\cite{mittal2013making} to quantitatively compare the methods. The NIQE score better suits unconstrained environments such as ours, as it measures deviations from natural image statistics, rather than tuning to specific distortions by training.
Each example displays the NIQE scores of the processed images, normalized to the NIQE score of the high-quality image found. As the NIQE score \textit{decreases} as quality \textit{increases}, \textbf{the closer the score is to 1 - the better the quality}.

It can be easily seen for all examples, both visually and quantitatively, that our algorithm yields better results than either the input dark image, the brightened noisy image or the BM3D denoised one. It does not only remove noise, but also embeds new HQ details, while preserving pose, expression and identity. Table \ref{tabular:NIQE} and Fig. \ref{fig::NIQE_graph_17} show that it outperforms the prior-based brightening method, which outperforms the input; BM3D processing is always worse then the input (one should note though, that using BM3D in such a blind model is quite far from the standard denoising model of AWGN with known noise variance, where BM3D performs very well).

\begin{table}[!h]
	\begin{center}
		\caption{normalized niqe quality assessment for different methods}
		\begin{tabular}{|
				>{\centering\arraybackslash}m{0.7in} |
				>{\centering\arraybackslash}m{0.4in} | >{\centering\arraybackslash}m{0.6in} | >{\centering\arraybackslash}m{0.4in} | >{\centering\arraybackslash}m{0.5in} |}
			\hline
			\textbf{    } & \textbf{Proposed method} & \textbf{ Prior-based brightened image} & \textbf{Input image} & \textbf{BM3D Denoising}
			%			\parbox{0pt}{\rule{0pt}{0ex+\baselineskip}}\\
			\\[0ex] \hline
			Average score & \textbf{1.1729} & 1.2512 & 1.3708 & 1.9056\\[0ex] \hline
			Relative imp. over input [\%] & \textbf{14.43} & 8.72 & 0 & -39.02 \\[0ex] \hline
		\end{tabular}	
		\label{tabular:NIQE}
	\end{center}
\end{table}

Difference images between prior-based brightened input images and our results (Figs. \ref{fig::denoising_142152}, \ref{fig::denoising_152347}) show how using personal priors not only removes noise, but also embeds image details and fine textures, e.g. in the eyes, eyebrows and mouth.
Fig. \ref{fig::denoising_142152} also shows a close-up comparison of significant facial regions (that attract human attention and convey facial expression) for different methods.
In addition, it displays the high-quality example images used to extract prior information: the image used to extract mouth \& head/skin information and for background illumination adjustment; and the image used to extract eyes information. Note that the mouth/head example is similar in pose and mouth expression to the input, but different in eye expression, background, hair, clothes, etc.

In Fig. \ref{fig::denoising_152347}, \ref{fig::errors_110038} we discuss the effect of errors or omission of certain stages in the algorithm. Fig. \ref{fig::denoising_152347} shows the necessity of the head registration and blending stages for visually reasonable results.
Fig. \ref{fig::errors_110038} shows the effect of erroneous example facial feature selection (relating to the example in Fig. \ref{fig::flow_algorithm}). {The Demon measure allows accurate selection. But what if the selection process resulted in errors?}
These could have been caused, for instance, by insufficient expression variations in the dataset; or when skipping the illumination adjustment phase (see Fig. \ref{fig::color_effect}).
Demon registration's \textbf{robustness to moderate non-rigid variations} allows it to {overcome moderate selection errors}, such that the resulting interpolated features are of quite desirable shape, but somewhat distorted. However, features interpolated using very wrongly selected examples display wrong and distorted expressions. Note the wrong nose and wrinkles when wrongly selecting the mouth expression.

\makeatletter
\setlength{\@fptop}{0pt}
\makeatother

\section{Conclusion}
In this work we aim to overcome classical image processing limits by combining semantic patches and registration methods for visual image enhancement.
We demonstrate our method for the problem of cellular photography enhancement of dark facial images.
Given today's easily available photography devices, our model assumes that high-quality personal priors are available, but that we are blind to the degradation model and its parameters. A low-to-moderate degradation may include an unknown mix of noise, nonlinear post-processing artifacts, certain motion blur, resolution reduction and color-change. The blind model assumption allows a very general correction mechanism which is not device and scenario dependent.
In order to solve this we use non-rigid semantic patches and a registration algorithm, which is robust to noise and blur, and can infer a high quality solution based on the priors.

The experimental results demonstrate how our method achieves significant quality enhancement over the degraded input images, both visually and quantitatively, based on the no-reference NIQE measure. %for different identities, expressions and poses.
Our building blocks are facial features of coherent structure and context with adaptive size and location. % non-rigid regions of  as ,
%while preserving their structural coherency.
%We define
A new affinity measure is defined based on the non-rigid, diffusion-based Demon registration.
%which is of high correspondence to real-world appearance of images interpolated during diffusion.
We use it to construct data-driven, high-quality facial features spaces, representing various expression variations.

The measure's robustness to image quality degradation and non-rigid variations allows accurate matches of low-quality features to high-quality examples. This enables high enhancement quality, relying on only tens of personal priors, maintaining well the person's features and facial expressions. In a future work we consider processing of more abstract non-facial data within a generalized framework.
% or


\appendices
\section{Demon Registration} \label{Demon Registration}
%{\color{red} Appendices are written well, did not find any problem here..}
The Demon registration is a diffusion-based image registration algorithm, approximating fluid registration, based on polarity. It uses the transformation field caused by edge-based forces. It was first introduced by Thirion~\cite{thirion1995fast},~‎\cite{thirion1998image} as an analogy of Maxwell's "Demons" in a paradox of thermodynamics.

\begin{figure}[!h]
	\captionsetup{justification=centering}
	\centering
	\includegraphics[width=3in]{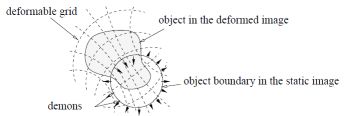}
	\caption{Demon diffusion process~\cite{thirion1998image}:  the "moving" object, represented by a deformable grid, diffuses through a semi-permeable membrane representing the boundaries of the "static" object, by the action of Demon effectors.}
	\label{fig::membrane}
\end{figure}

Fig. \ref{fig::membrane}, taken from Thirion's work~\cite{thirion1998image}, shows the Demon diffusion process. An object in the deforming image, referred to as "the moving image", is represented by a deformable grid, whose nodes are labeled "inside" or "outside"; their inner relations correspond to object rigidity. The boundaries of an object in the other image, referred to as "the static image", are represented by a semi-permeable membrane, along which Demon effectors are situated.
The deformable grid gradually diffuses into the static object through its boundaries by the action of these effectors. Diffusion is guided by the principle of polarity, that is, maximal common polarity at each side of the membrane: Demon effectors act to locally "push" nodes labeled "inside" through the membrane interface into the static object, and vice versa. To this end, Demons might use spatial location, direction, pixel intensity or other information.

The final transformation results from iteratively evolving a family of transforms under two types of forces: "internal" forces, reflecting inner relations between neighboring image points, corresponding to image rigidity; and "external" forces, reflecting interaction between the static and moving images.

\begin{figure}[!t]
	\centering
	\captionsetup{justification=centering}
	\subfloat[From left to right: original image and intermediate images generated for 200, 400, 500 and 700 iterations.]{\includegraphics[width=3in]{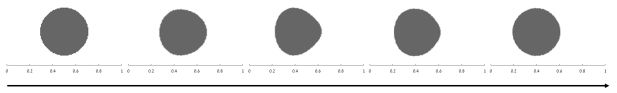}%
		\label{fig::img_full_conv}}
	\hfil
	\subfloat[Mean absolute error between deformed and target images vs. number of iterations]{\includegraphics[width=1.5in]{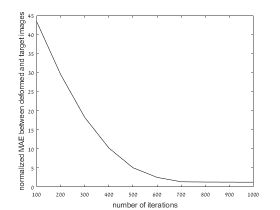}%
		\label{fig::graph_full_conv}}
	\caption{Intermediate steps in the diffusion process for object translation.}
	\label{fig::full_converge}
\end{figure}

\begin{figure}[!t]
	\captionsetup{justification=centering}
	\centering
	\subfloat[From circle to square.]{\includegraphics[width=1.5in]{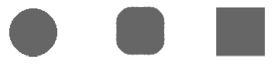}%
		\label{fig::circle2square}}
	\qquad
	\subfloat[From square to circle.]{\includegraphics[width=1.5in]{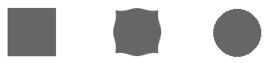}%
		\label{fig::square2circle}}
	\caption{Deformation of a circle to a square, and vice versa (for 200 iterations). Left: source image. Middle: deformed image. Right: target image.}
	\label{fig::circle_square}
\end{figure}

Fig. \ref{fig::full_converge} illustrates this, showing the intermediate steps of diffusing an object into a same-shape translated object, until perfect registration is achieved (Fig. \ref{fig::img_full_conv}), and the mean absolute error between deformed and target images as a function of the number of iterations (Fig. \ref{fig::graph_full_conv}). Fig. \ref{fig::circle_square} demonstrates the difficulty in deforming a high-curvature shape into a low-curvature shape \textit{in a given time}, and vice versa.

Thirion~\cite{thirion1995fast} showed the translation of this concept into a simple gradient-based displacement field  $\vec{u}$, to estimate the displacement of a pixel in the \textit{moving image} $m$, required to match the corresponding point in the \textit{static image} $s$.‎

%Denote $\vec{u}$ as the displacement vector which brings the \textit{moving image} $m$ closer to the \textit{static image} $s$.
Denoting pixel intensity as a function of time: $i(x(t),y(t),z(t),t)$, differentiating the instantaneous optical flow equation gives:
\begin{equation}
\frac{\partial i}{\partial x}\frac{\partial x}{\partial t} + \frac{\partial i}{\partial y}\frac{\partial y}{\partial t}+ \frac{\partial i}{\partial z}\frac{\partial z}{\partial t}= -\frac{\partial i}{\partial t}
\end{equation}
Considering that the evolution in one time unit is the difference between images: $\frac{\partial i}{\partial t}=s-m$, and $\vec{u}=(\frac{dx}{dt},\frac{dy}{dt},\frac{dz}{dt})$ is the instantaneous velocity from $m$ to $s$, we get:
\vspace{-0.1cm}
\begin{equation}
\vec{u} \cdot  \vec\nabla s= m-s
\end{equation}
where $\vec\nabla$ denotes image gradient. Defining $\vec\nabla s$ as the internal edge-based force, and $(m-s)$ as the external force, $\vec{u}$ is computed locally as the shortest translation of a point of $m$ onto the hyperplane approximating $s$:
\begin{equation}
\vec{u} = \frac {(m-s)\vec\nabla s}{|\vec\nabla s|^2}
\end{equation}
%Where $\vec\nabla s$ is the internal edge-based force, and $(m-s)$ is the external force.
Unfortunately, small intensity variations can result in infinite Demon forces. To stabilize the resulting unstable equation:
%, which makes this equation unstable. To avoid that, the equation is stabilized by:
\begin{equation}
\vec{u} = \frac {(m-s)\vec\nabla s}{|\vec\nabla s|^2 + (m-s)^2}
\end{equation}
To improve stability and convergence speed, Wang \textit{et al}.~‎\cite{wang2005validation} added an "active force". Diffusion was considered a bi-directional process, and therefore Demon effectors also produced an internal gradient-based force of $m$, that diffuses $s$ into $m$. A normalization factor $\alpha$ is used to account for the adaptive force strength adjustment (suggested by Cachier \textit{et al}.~‎\cite{cachier1999fast}), yielding the following displacement field:
\begin{equation}
\vec{u} = (m-s) \times \left (\frac {\vec\nabla s}{|\vec\nabla s|^2 + \alpha^2 (s-m)^2}+\frac {\vec\nabla m}{|\vec\nabla m|^2 + \alpha^2 (s-m)^2}\right)\label{Demon}
\end{equation}
The simple, iterative Demon registration algorithm introduced by Wang \textit{et al}. consists of the following steps:
\begin{enumerate}
	\item Calculation of the disp. field using Eq. \eqref{Demon}.
	\item Regularization of the disp. field using Gaussian smoothing, to suppress noise and preserve geometric continuity. %of $m$.
	\item Adding the regularized disp. field to the total disp. field.
	\item Image deformation according to the total disp. field.
\end{enumerate}
Cachier \textit{et al}. showed in their work that the Demon algorithm can be seen as an approximation of a second order gradient descent of a SSD criterion, and proposed using this criterion to compare different non-rigid registration methods. But, as opposed to our work, it was not used as an affinity criterion or to evaluate the success of image deformation.

This registration method was so far usually used for medical image registration, such as the work of Kroon and Slump~\cite{kroon2009mri}.

Another non-rigid registration, penalizing mismatch between deformed and target images, is optimal mass transport, finding the cheapest mass transport path, by minimizing the L2 Kantorovich-Wasserstein distance. Benamou and Brenier~\cite{benamou2000computational} proved it can be \textit{reformulated as a fluid mechanics problem}. 

%Though it may be possible to use optimal transport or other fluid methods in a framework similar to ours, as far as we know, it hasn't been done yet. But, some existing works do relate to ours.

As far as we know, optimal transport or other fluid methods haven't been used so far in a framework similar to ours; but, some existing works do relate to ours. Haker \textit{et al}.~\cite{haker2004optimal} used OT for image registration, aiming to achieve the minimal possible distance, while interpolating intermediate images (interpolation was also done by Kerrache and Nakauchi~\cite{kerrache2011interpolation}). This is as opposed to our \textit{time-limited} process aimed to determine \textit{visual validity}-related affinities between images. Wang \textit{et al}.~\cite{wang2013linear} used a \textit{linear} approximated framework to quantify and visualize variations in a set of images; but their image approximation becomes less accurate as images become less sparse, and might not suit our needs.
% On the other hand, avoiding these approximations will increase running time drastically.
Kolouri and Rohde~\cite{kolouri2015transport} represented displacement fields between \textit{multiple-identity} HR facial images as \textit{linear} combinations of basis fields, to constrain a super-resolution procedure. And last, Hassanien and Nakajima~\cite{hassanien1998image} used the PDE-based Navier elastic body splines for morphing and interpolation between \textit{multiple-identity} facial images, but their method requires knowledge of feature points and their correspondences.

\section{Valid and Non-valid Demon deformation examples}\label{synthetic}
\begin{figure}[!t]
	\captionsetup{justification=centering}
	\centering
	\subfloat[Scale factor=0.5]{\includegraphics[width=1in]{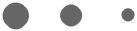}%
		\label{fig::scale1}}
	\qquad	
	\subfloat[Scale factor=0.8]{\includegraphics[width=1in]{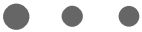}%
		\label{fig::scale2}}
	\\
	\subfloat[Scale factor=1.2]{\includegraphics[width=1in]{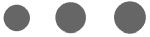}%
		\label{fig::scale3}}
	\qquad	
	\subfloat[Scale factor=1.5]{\includegraphics[width=1in]{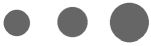}%
		\label{fig::scale4}}
	\\	
	\subfloat[Demon distance as a function of scale factor]{\includegraphics[width=1.5in]{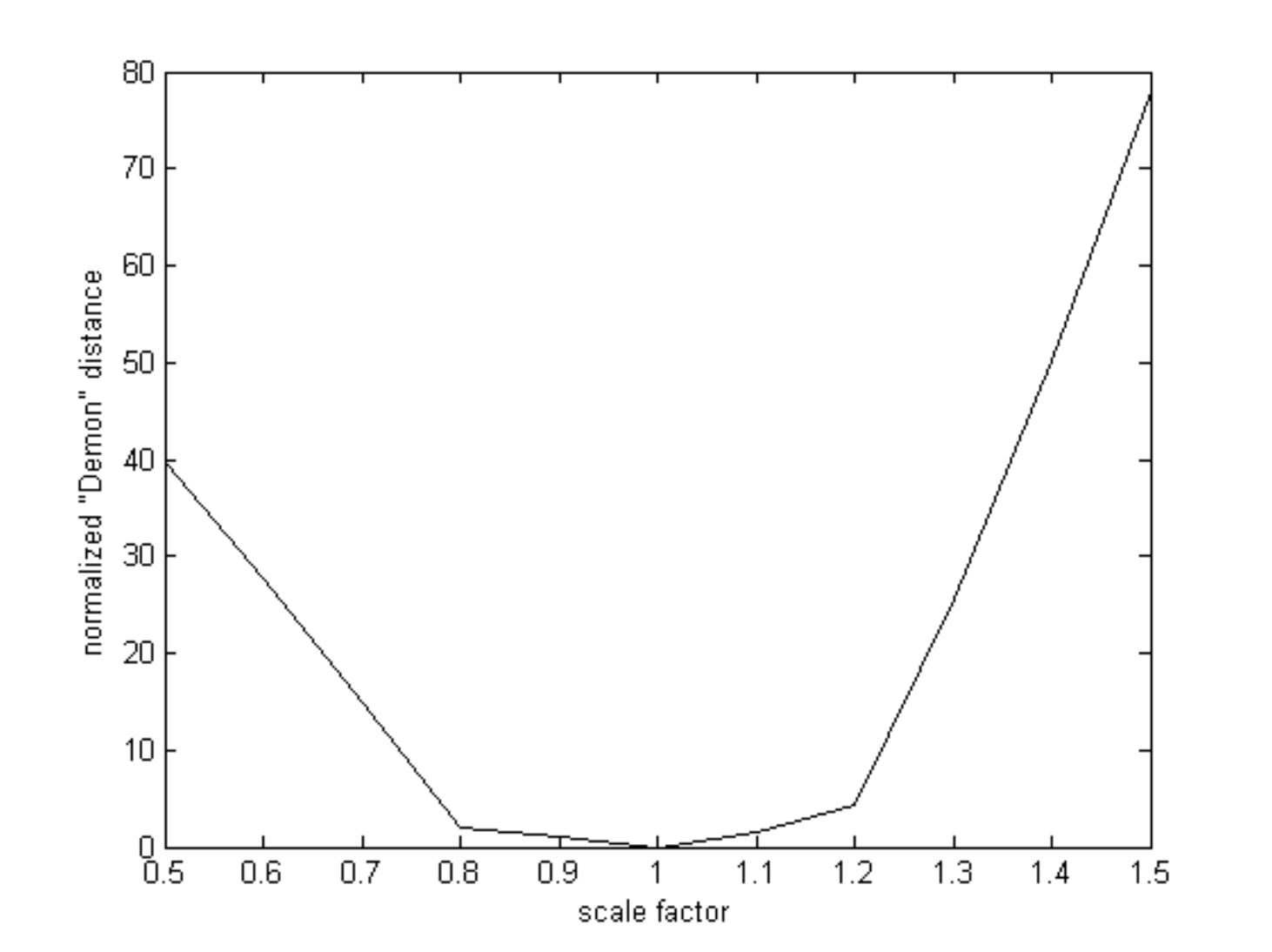}%{scaling_graph.eps}%
		\label{fig::scaling_graph}}	
	\caption{Demon deformation for object scaling at different factors. For each scale factor (a)-(d): Left: source image. Middle: deformed image. Right: target image, that is, the source image scaled.}
	\label{fig::scaling}
\end{figure}

\begin{figure}[!t]
	\captionsetup{justification=centering}
	\centering
	\subfloat[Translation=-15 pixels]{\includegraphics[width=1.2in]{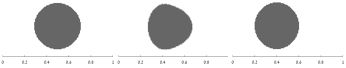}%
		\label{fig::trans1}}
	\qquad
	\subfloat[Translation=-10 pixels]{\includegraphics[width=1.2in]{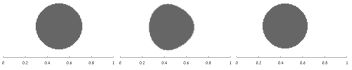}%
		\label{fig::trans2}}
	\\
	\subfloat[Translation=-8 pixels]{\includegraphics[width=1.2in]{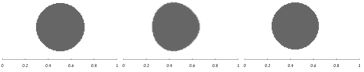}%
		\label{fig::trans3}}
	\qquad
	\subfloat[Translation=-5 pixels]{\includegraphics[width=1.2in]{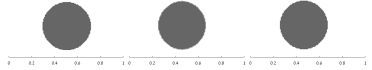}%
		\label{fig::trans4}}
	\\
	\subfloat[Demon distance as a function of translation]{\includegraphics[width=1.5in]{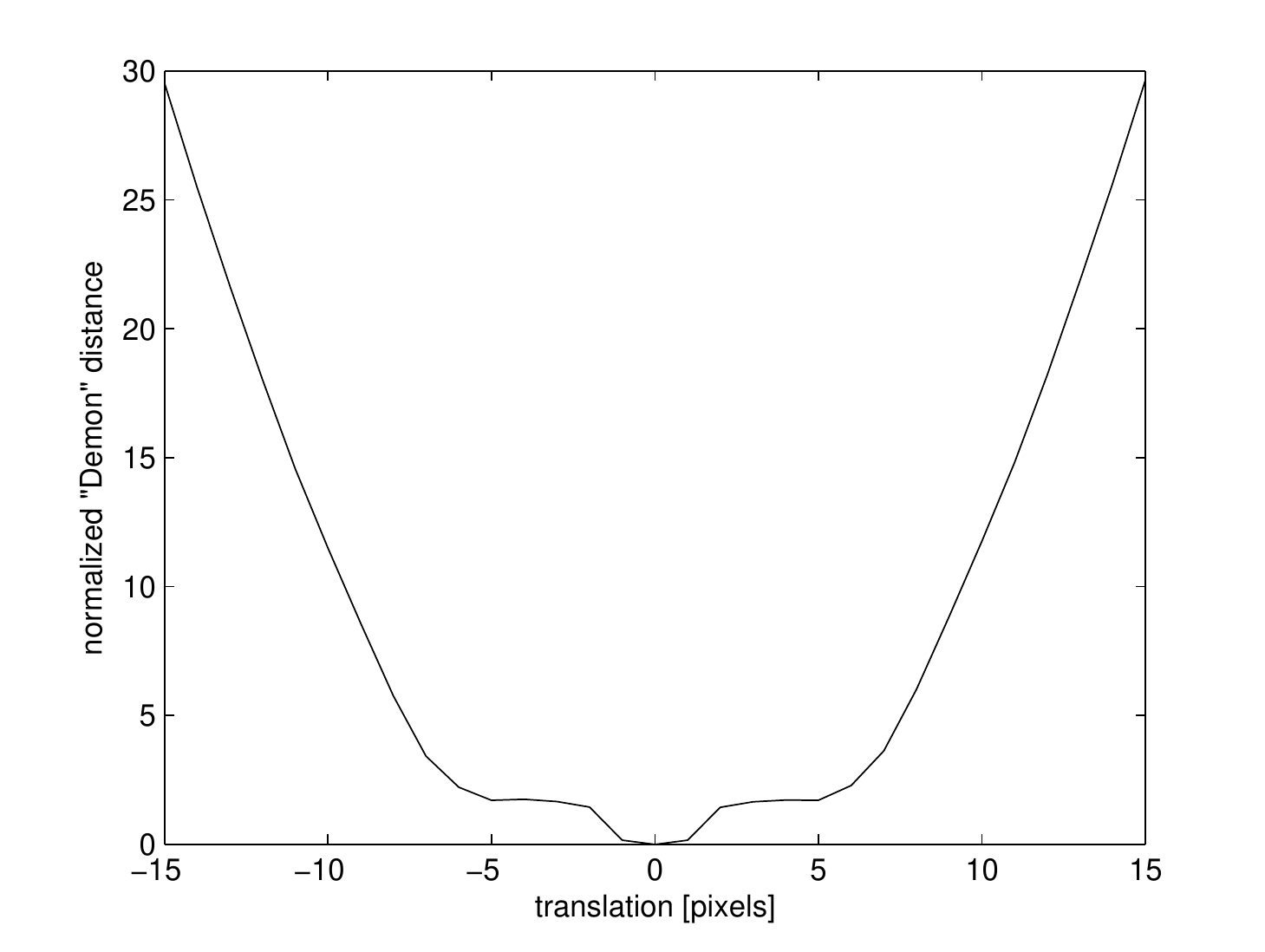}%{trans_graph.eps}%
		\label{fig::trans_graph}}
	\caption{Demon deformation for different object translations. For each translation (a)-(d): Left: source image. Middle: deformed image. Right: target image, that is, the source image translated.}
	\label{fig::translation}
\end{figure}

\begin{figure}[!t]
	\captionsetup{justification=centering}
	\centering
	\subfloat[Rotation angle=30$^{\circ}$]{\includegraphics[width=1.2in]{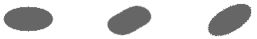}%
		\label{fig::rot1}}
	\qquad	
	\subfloat[Rotation angle=20$^{\circ}$]{\includegraphics[width=1.2in]{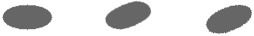}%
		\label{fig::rot2}}
	\\
	\subfloat[Rotation angle=14$^{\circ}$]{\includegraphics[width=1.2in]{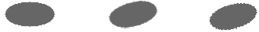}%
		\label{fig::rot3}}
	\qquad
	\subfloat[Rotation angle=4$^{\circ}$]{\includegraphics[width=1.2in]{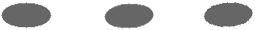}%
		\label{fig::rot4}}
	\\
	\subfloat[Demon distance as a function of rotation angle]{\includegraphics[width=1.5in]{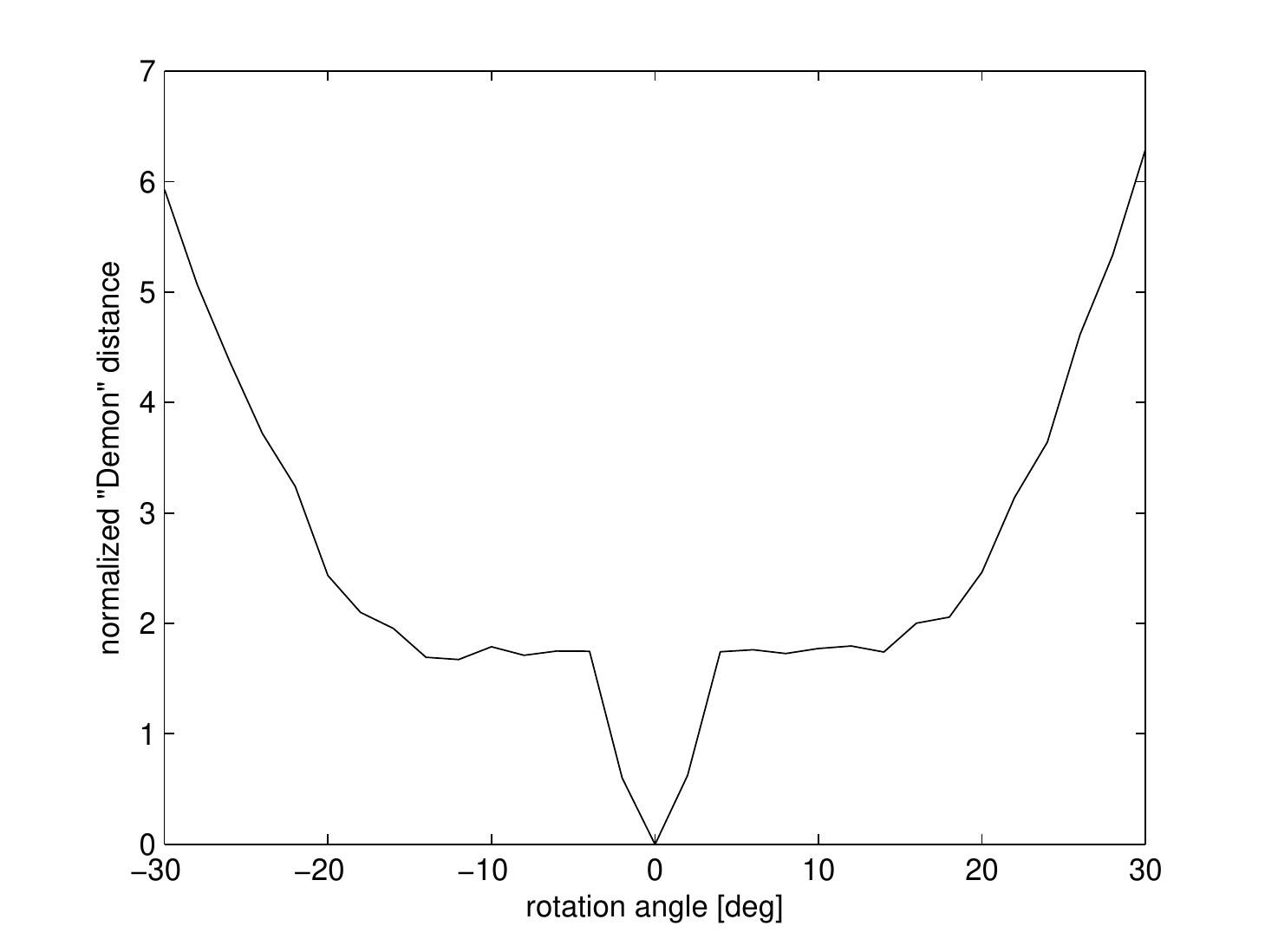}%{rot_graph.eps}%
		\label{fig::rot_graph}}
	\caption{Demon deformation for object rotation at different angles. For each rotation angle (a)-(d): Left: source image. Middle: deformed image. Right: target image, that is, the source image rotated.}
	\label{fig::rotation}
\end{figure}

We demonstrate the behavior of \textit{time-limited} Demon deformation and measure (using 200 iterations), under basic affine transformations. Fig. \ref{fig::scaling} demonstrates the \textit{visual} success of the deformation and measure for different scale variations. Note, that changing scale involves the dis/appearance of mass. Similarly, Fig. \ref{fig::translation} and \ref{fig::rotation} display the behavior for different translations or rotation angles, respectively.

These experiments all illustrate the same behavior: for moderate variations the \textit{time-limited} Demon deformation succeeds: the deformed image gets as close as possible to the target image (practically identical) (Figs. \ref{fig::scale2}, \ref{fig::scale3}, \ref{fig::trans4}, \ref{fig::rot3}, \ref{fig::rot4}); and the Demon measure \textit{moderately} increases with variation.
But there exists a breaking point where the \textit{time-limited} deformation starts to fail: the deformed image is too different from the target image, or distorted (Figs. \ref{fig::scale1}, \ref{fig::scale4}, \ref{fig::trans1} to \ref{fig::trans3}, \ref{fig::rot1}, \ref{fig::rot2}); and the measure starts to \textit{drastically} increase.

%For example, moderate scale variations (Figs. \ref{fig::scale2}, \ref{fig::scale3}) as opposed to more drastic scale variations (Figs. \ref{fig::scale1}, \ref{fig::scale4}); Medium-small extents of translation (Fig. \ref{fig::trans4}), as opposed to larger translations (Figs. \ref{fig::trans1} to \ref{fig::trans3}), where the deformed image becomes distorted; And successful deformation at medium-small rotation angles (Figs. \ref{fig::rot3}, \ref{fig::rot4}), as opposed to distorted deformation images at greater rotation angles (Figs. \ref{fig::rot1}, \ref{fig::rot2}).
\begin{figure}[!t]
	\captionsetup{justification=centering}
	\centering
	\subfloat[Eye gaze=-15 pixels]{\includegraphics[width=1.2in]{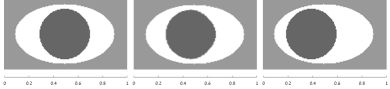}%
		\label{fig::gaze1}}
	\qquad
	\subfloat[Eye gaze=-7 pixels]{\includegraphics[width=1.2in]{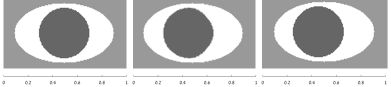}%
		\label{fig::gaze2}}	
	\\
	\subfloat[Eye gaze=-5 pixels]{\includegraphics[width=1.2in]{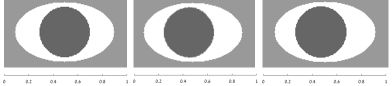}%
		\label{fig::gaze3}}
	\qquad
	\subfloat[Eye gaze=5 pixels]{\includegraphics[width=1.2in]{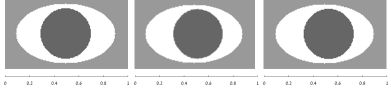}%
		\label{fig::gaze4}}
	\\
	\subfloat[Demon distance and source to target MAE as a function of eye gaze translation]{\includegraphics[width=1.5in]{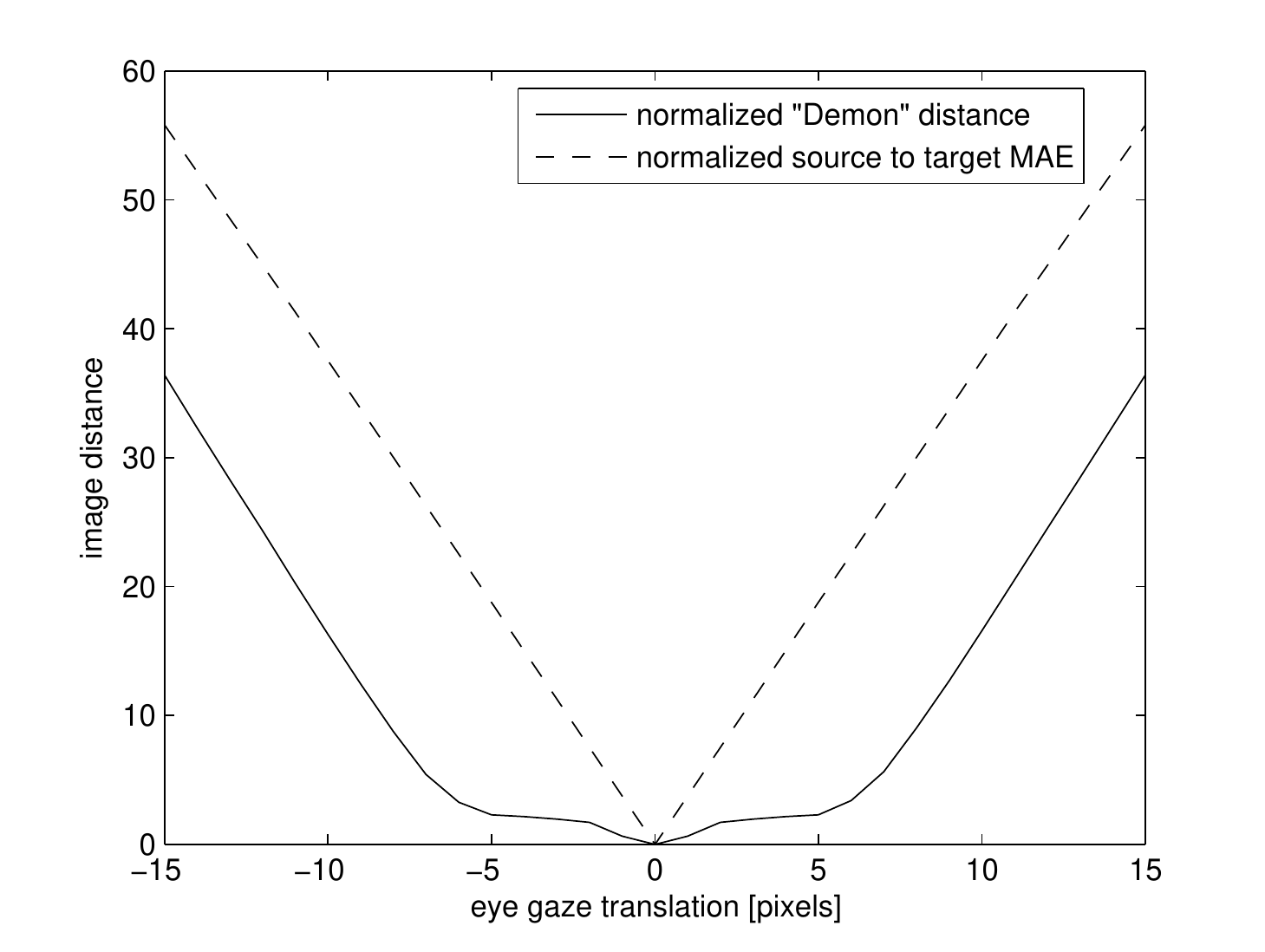}%{eye_gaze_graph.eps}%
		\label{fig::eye_gaze_graph}}
	\caption{Demon deformation for different eye gaze translations. For each translation (a)-(d): Left: source image, depicting a central gaze. Middle: deformed image. Right: target image, depicting a gaze change.}
	\label{fig::eye_gaze}
\end{figure}

We also explore this behavior for a common non-rigid facial expression deformation: a change in eye gaze (Fig. \ref{fig::eye_gaze}).
As seen before, moderate variations in eye gaze result in successful deformations and a moderate increase in Demon distance (Figs. \ref{fig::gaze3}, \ref{fig::gaze4}), whereas greater variations result in \textit{distorted} images and a drastic increase in distance (Figs. \ref{fig::gaze1}, \ref{fig::gaze2}).
Fig. \ref{fig::eye_gaze_graph} compares the same breaking-point behavior seen before, to the \textit{linear} behavior of the MAE between \textit{source} and target images, not reflecting the deformation's success.
% give a sense of how natural or distorted the resulting interpolated images are. %We will later address this concept of image distortion as the visual validity of the deformation between images.
\section*{Acknowledgment}
The authors would like to thank Yossi Bar Erez, Raz Nossek and Alona Baruhov for their assistance.

%\ifCLASSOPTIONcaptionsoff
%  \newpage
%\fi



%\bibliographystyle{IEEEtran}
\bibliography{DemonBib}{}
\bibliographystyle{IEEEtran}
%\begin{thebibliography}{1}
%\end{thebibliography}

\vspace*{-1.55 cm}
\begin{IEEEbiographynophoto}{Ester Hait}
	received her B.Sc. degree in Electrical Engineering (Cum Laude) from the Technion, Haifa, Israel, in 2014. She is currently pursuing her M.Sc. degree in Electrical Engineering in the Technion. Her research interests include image processing and computer vision.
\end{IEEEbiographynophoto}
\vspace*{-1 cm}
%\begin{IEEEbiography}
\begin{IEEEbiographynophoto}{Guy Gilboa}
	 is a faculty member at the Department of Electrical Engineering, Technion - Israel Institute of Technology, since 2013.
\end{IEEEbiographynophoto}


%\vfill


\end{document}